\documentclass[11pt,a4paper,acceptedWithA]{article}
\usepackage{times,latexsym}
\usepackage{url}
\usepackage[T1]{fontenc}
\usepackage{tipa}
\usepackage{amsmath}
\usepackage{amsfonts}
\usepackage{amssymb}
\usepackage{mathrsfs}
\usepackage{graphicx}
\usepackage[]{hyperref}
\usepackage[american]{babel}
\usepackage{multirow}
\usepackage{pifont}
\usepackage{float}
\usepackage{natbib}
\usepackage{tipa}
\usepackage{tikz}

\usepackage[acceptedWithA]{tacl2018v2}

\usepackage{xspace,mfirstuc,tabulary}

\newif\iftaclinstructions
\taclinstructionsfalse 
\iftaclinstructions
\renewcommand{\confidential}{}
\renewcommand{\anonsubtext}{(No author info supplied here, for consistency with
TACL-submission anonymization requirements)}
\newcommand{\instr}
\fi

\iftaclpubformat 

\else

\fi

\title{Identity-Based Patterns in Deep Convolutional Networks: Generative Adversarial Phonology and  Reduplication}

\author{
 Ga\v{s}per Begu\v{s}
 \\
 University of California, Berkeley \\
  {\sf begus@berkeley.edu} \\
}

\date{}

\begin{document}
\maketitle

\begin{abstract}

This paper models unsupervised learning of an identity-based pattern (or copying) in speech called reduplication from  raw continuous data with deep convolutional neural networks. We use the ciwGAN architecture \citep{begusCiw} in which learning of meaningful representations in speech emerges from a requirement that the CNNs generate informative data. We propose a technique to wug-test CNNs trained on speech and, based on four generative tests,  argue that the network learns to represent an identity-based pattern in its latent space. By manipulating only two categorical variables in the latent space, we can actively turn an unreduplicated form into a reduplicated form with no other substantial changes to the output in the majority of cases.  We also argue that the network extends the identity-based pattern to unobserved data. Exploration of how meaningful representations of identity-based patterns emerge in CNNs and how the latent space variables outside  of the training range correlate with identity-based patterns in the output has general implications for neural network interpretability.

\end{abstract}

\section{\label{intro} Introduction}

The relationship between symbolic representations and connectionism has been subject to ongoing discussions in computational cognitive science. Phonology offers a unique testing ground in this debate as it is concerned with the first discretization that human language users perform: from continuous phonetic data to discretized mental representations of meaning-distinguishing sounds called \emph{phonemes}.

Identity-based patterns, repetition, or copying have been at the center of this debate \citep{marcus99}. Reduplication is a morphophonological process where phonological content (phonemes) get copied from a word (also called the base) with some added meaning \citep{inkelas05,urbanczyk17}. It can be total, which means that all phonemes in a  word get copied (e.g.~/pula/ $\rightarrow$ [\textbf{pula}-pula]), or partial, where only a subset of segments gets copied (e.g.~/\text{pula}/ $\rightarrow$ [\textbf{pu}-pula]). 

Reduplication is indeed unique among  processes in natural language because combining learned entities based on training data distributions does not yield the desired outputs. For example, a learner can be presented with pairs of bare and reduplicated words, such as /pala/ $\sim$ /papala/ and /tala/ $\sim$ /tatala/. The learner can then be tested on providing a reduplicated variant of a novel unobserved item with an initial sound /k/ that they have not been exposed to (e.g.~/kala/). If the learner learns the reduplication pattern, they will output [kakala].  If the learner simply learns that /pa/ and /ta/ are optional constituents that can be attached to words based on data distribution, they will output [pakala] or [takala]. Reduplication is thus an identity-based pattern (similar to copying), which is computationally more challenging to learn \citep{gasser93}, both in connectionist \citep{brugiapaglia20} and non-connectionist frameworks \cite{savitch89,dolatian20}. In /k$_i$a$_j$k$_i$a$_j$la/, the two sounds in the reduplicative morpheme, /k$_i$/ and /a$_j$/, need to be in an identity relationship with the first two segments of the base, /k$_i$/ and /a$_j$/, and the learner needs to copy rather than recombine learned elements.

 \citet{marcus99} argue that connectionist models such as simple neural networks are unable to learn a simple reduplication pattern that 7-month old human infants are able to learn (see also \citealt{gasser93}).  According to \citet{marcus99},  the behavioral outcomes of their experiments cannot be modeled by simple counting, attention to statistical trends in the input, attention to repetition, or connectionist (simple neural network) computational models. Instead, they argue, the results support the claim that human infants employ ``algebraic rules'' \citep{marcus99,marcus01,berent13} to learn reduplication patterns (for a discussion, see, among others, \citealt{mcclelland99,endress07}).

With the development of neural network architectures, several studies revisited the claim that neural networks are unable to learn reduplicative patterns \citep{alhama18,prickett18,nelson20,brugiapaglia20}, arguing that identity-based patterns can indeed be learned with more complex architectures.\footnote{\citet{wilson18,wilson20} proposes another approach that allows modeling reduplication. For a non-connectionist computational model of reduplication, see \citet{dolatian18,dolatian20}.} All these computational experiments, however,  operate on an already discretized level and most of these experiments model reduplication with supervised learning. 

Examples like [\textbf{pu}-pula] and [\textbf{pula}-pula] are discretized representations of reduplication, using characters to represent continuous sounds. Most, if not all, computational models of reduplication, to the author's knowledge, model reduplication as character or feature manipulation (the inputs to the models are either characters representing phones or phonemes or discrete binary featural representations of phonemes). For example, a seq2seq model treats reduplication as a pairing between the input unreduplicated sequence of ``characters'' (such as /tala/) and an output --- a reduplicated sequence (such as /tatala/).
Already abstracted and discretized phonemes or ``characters'', however, are not the primary input to language-learning infants. The primary linguistic data of most hearing infants is raw continuous speech. Hearing infant learners need to acquire reduplication from continuous speech data that is substantially more complex than already discretized characters or binary features.

Furthermore, most of the existing models of reduplication are also supervised. 
Seq2seq models, for example, are fully supervised: the training consists of pairs of unreduplicated (input) and reduplicated strings of characters or binary features (output). While the performance can be tested on unobserved data or even on unobserved segments, the training is nevertheless supervised. Human language learners do not have access to input-output pairings: they are only presented with positive, surface,  and continuous acoustic data. While equivalents of copying/identity-based patterns can be constructed in the visual domain, we are not aware of studies that test identity-based visual patterns with deep  convolutional neural networks.

 In this paper, we model reduplication, one of the computationally most challenging processes, from raw unlabeled acoustic data with deep convolutional networks in the GAN framework. The advantage of the GAN framework for cognitive modeling is that the network has to learn to output raw acoustic data from a latent noise distribution without directly accessing the training data. We argue that CNNs discretize continuous phonetic data and encode linguistically meaningful units into individual latent variables. The emergence of a discretized representation of an identity-based pattern (reduplication) is induced by a model which forces the Generator network to output informative data (ciwGAN; Section \ref{CiwGAN}). Additionally, we add inductive bias towards symbolic-like representations by binarizing code variables with which the Generator encodes meaningful representations. We also test whether a deep convolutional network learns reduplication without the two inductive biases (without the requirement on the Generator to output informative data and without binarization of the latent space) in the bare WaveGAN architecture (Section \ref{reproduction}). 
 
 The experiments bear implications for the discussion between symbolic and connectionist approaches to language modelling by testing the emergence of rule-like symbolic representations within the connectionist framework from raw speech in an unsupervised manner. Results of the experiments suggest that both models, ciwGAN and WaveGAN learn the identity-based patterns, but inductive biases for informative representation and binarization facilitate learning and yield better results.
We discuss properties of symbolic-like representations and how they emerge in the models: discretization, causality (in the sense that manipulation of individual elements results in desired outcome), and categoricity (for discussions on these and other aspects of the debate, see \citealt{rumelhart86,mcclelland86,fodor88,minsky91,dyer91,marcus99,marcus01,manning03,berent13,maruyama21}).

How can we test learning of reduplication in a deep convolutional network that is trained only on raw positive data? We propose a technique to test the ability of the Generator to produce forms absent from the training data set. For example,  we train the networks on acoustic data of items such as /pala/ $\sim$ /papala/ and /tala/ $\sim$ /tatala/, but test reduplication on acoustic forms of items such as /sala/, which is never reduplicated in the training data.  Using the technique proposed in \citet{begus19}, we can identify latent variables that correspond  to some phonetic or phonological representation such as reduplication. By manipulating and interpolating a single latent variable, we can actively generate data with and without reduplication. In fact, we can observe a direct relationship between a single latent variable (out of 100) and reduplication that with interpolation gradually disappears from the output. Additionally, we can identify  latent variables that correspond to [s] in the output. By forcing both reduplication and [s] in the output through latent space manipulation, we can ``wug-test'' the network's learning of reduplication on unobserved data. In other words, we can observe what the network will output if we force it to output reduplication and an [s] at the same time. A comparison of generated outputs with human outputs that were withheld from training reveals a high degree of similarity. We perform an additional computational experiment to replicate the results from the first experiment (from Section \ref{CiwGAN}).  In the replication experiment, evidence for learning of the reduplicative pattern also emerges.  To the author's knowledge, this is the first attempt to model reduplication with neural network architectures trained on raw acoustic speech data.

The computational experiments reveal another property about representation learning in deep neural networks:
we argue that the network extracts information in the training data and represents a continuous acoustic identity-based pattern with discretized representation. Out of 100 variables, the network encodes reduplication with one or two variables, which is suggested by the fact that a small subset of variables are substantially more strongly correlated with presence of reduplication. In other words, there is a near categorical drop in regression estimates between one variable and the rest of the latent space. Setting the identified variables to values well beyond the training range results in near categorical presence of a desired variable in the output. This technique (proposed for non-identity-based patterns in \citealt{begus19}) allows us to directly explore how the networks encode dependencies in data, their underlying values, and interactions between variables, and thus get a better understanding of how exactly deep convolutional networks encode meaningful representations.

Recent developments in zero-resource speech modeling \cite{dunbar17,dunbar19,dunbar20} enable modeling of speech processes in an unsupervised manner from raw acoustic data. Several proposals exist for modeling unsupervised lexical learning  \cite{kamper14,lee15,chung16} that include generative models such as variational autoencoders \citep{chung16,baevski20,niekerk20} and GANs \cite{begusCiw}. This framework allows not only unsupervised lexical term discovery, but also phone-level identification \cite{eloff19,shain19,chung16,chorowski19}. While zero-resource speech modeling has yielded promising results in unsupervised labeling, the proposals generally do not model phonological or morphophonological processes.  This paper thus also tests applicability of the unsupervised speech processing framework for cognitive modeling and network interpretability.

\section{\label{reduplication} Model}

Generative Adversarial Networks \citep{goodfellow14} are a neural network architecture with two main components: the Generator network and the Discriminator network. The Generator is trained on generating data from some latent space that is randomly distributed. The Discriminator takes real training data and the Generator's outputs and estimates which inputs are real and which  are generated. The minimax training, where the Generator is trained on maximizing the Discriminator's error rate and the Discriminator is trained on minimizing its own error rate, results in the Generator network outputting data such that the Discriminator's success in distinguishing them  from real data is low. It has been shown that GANs not only learn to produce innovative data that resemble speech, but also learn to encode phonetic and phonological representations in the latent space \citep{begus19}. 
The major advantage of the GAN architecture for modeling speech is that the Generator network does not have direct access to the training data and is not trained on replicating data (unlike in the autoencoder architecture; \citealt{rasanen16,eloff19,shain19}). Instead, the network has to learn to generate data from noise in a completely unsupervised manner --- without ever directly accessing the training data.

In the first experiment, we use the ciwGAN (Categorical InfoWaveGAN) model proposed in \citet{begusCiw}. The ciwGAN model combines the WaveGAN and InfoGAN architectures. WaveGAN, proposed by \citet{donahue19}, is a Deep Convolutional Generative Adversarial Network (DCGAN; proposed by \citealt{radford15}) adapted for time-series audio data. The basic architecture is the same as in DCGAN, the main difference being that  in the WaveGAN proposal, the deep convolutional networks take one-dimensional time-series data as inputs or outputs. The structure of the Generator and the Discriminator networks in the ciwGAN architecture are taken from \citet{donahue19}. InfoGAN \cite{chen16} is an extension of the GAN architecture that aims to maximize mutual information between the latent space and generated outputs. The Discriminator/Q-network learns to retrieve the Generator's latent categorical or continuous codes \citep{chen16} in addition to estimating realness of generated outputs and real training data. 

\begin{figure*}
\centering
\includegraphics[width=.49\textwidth]{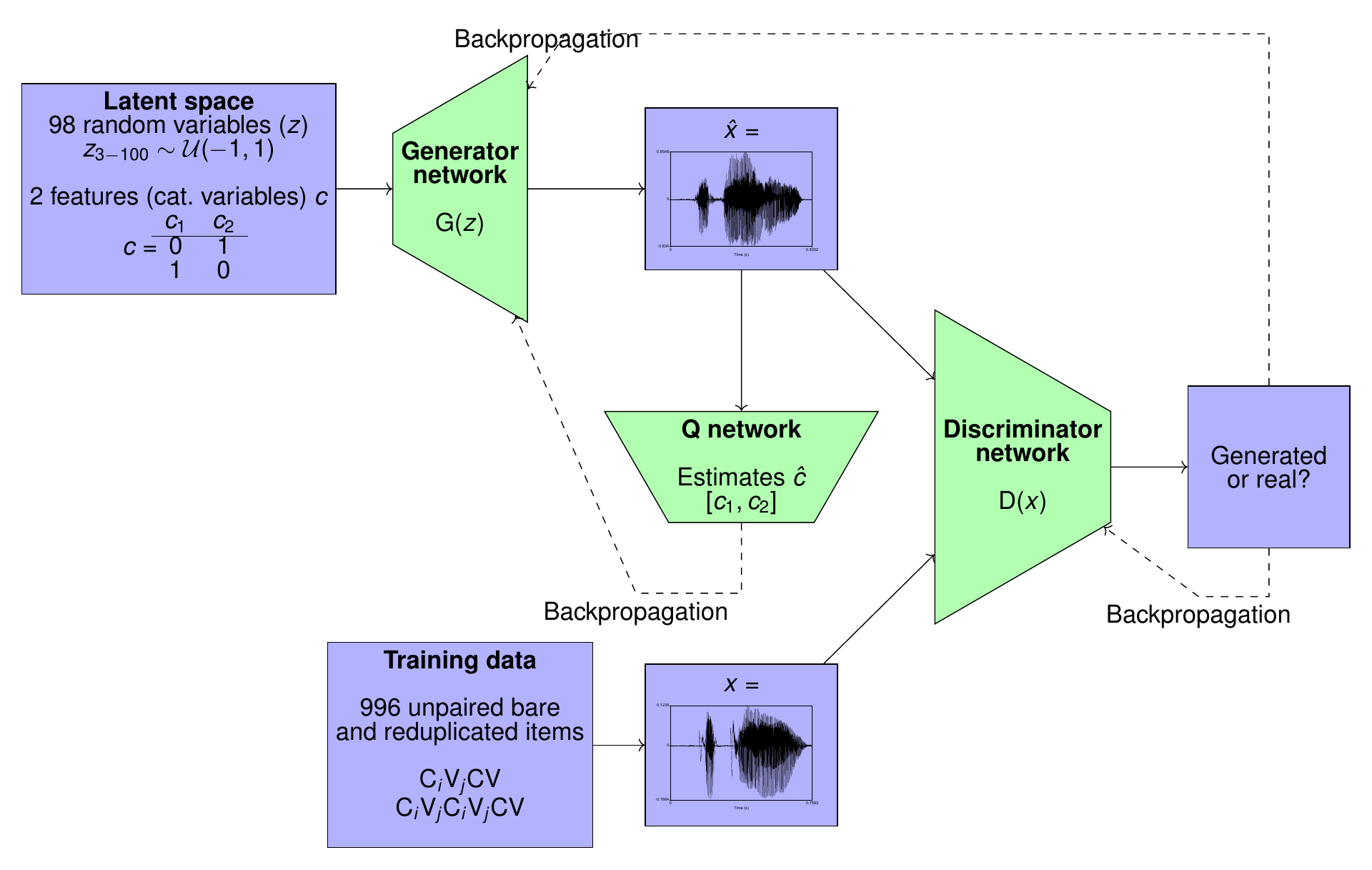} \includegraphics[width=.49\textwidth]{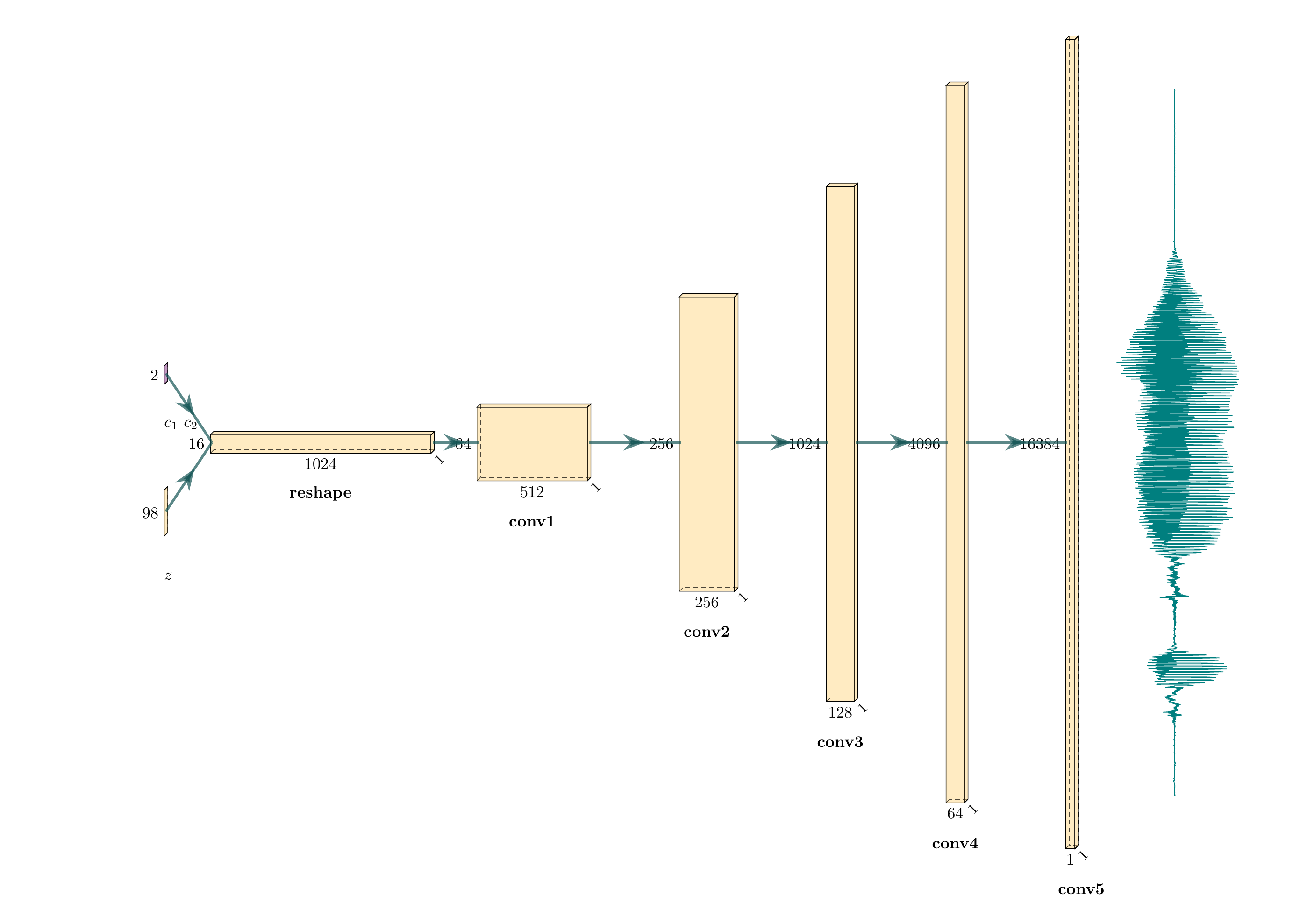}
\caption{\label{ciwRedgantikz} (left) The ciwGAN architecture as proposed in \citet{begusCiw} and used in this paper with training data as described in Section \ref{procedures}. (right) The structure of the Generator in the ciwGAN architecture as proposed in \citet{begusCiw} (based on \citealt{donahue19}). 
} 
\end{figure*}

\citet{begusCiw} proposes a model that combines these two proposals and introduces a new latent space structure (in the fiwGAN architecture). Because we are primarily interested in simple binary classification between bare and reduplicated forms, we use the ciwGAN variant of the proposal. The model introduces a separate deep convolutional Q-network that learns to retrieve the Generator's internal representations. Separating the Discriminator and the Q-network into two networks is advantageous from the cognitive modeling perspective: the architecture features a separate network that models speech production (the Generator) and a separate network that models speech categorization (the Q-network). The latter introduces an inductive bias  that forces the Generator to output informative data and encode linguistically meaningful properties into its code variables. The network learns to generate data such that by manipulating these code variables, we can force the desired linguistic property in the output \cite{begusCiw}.

The architecture involves three networks:  the Generator that takes latent codes (a one-hot vector) and uniformly distributed $z$-variables and generates waveforms, a Discriminator that distinguishes real from generated outputs, and a Q-network that takes generated outputs and estimates the latent code (one-hot vector) used by the Generator. More specifically, the Generator network is a deep convolutional network that takes as its input 100 latent variables (see Figure \ref{ciwRedgantikz}).\footnote{The number of latent variables were adopted from \citet{radford15} and \citet{donahue19}. Probing how the number of $z$-variables affects learning of speech representations is left for future work.} Two of the 100 variables are code variables ($c_1$ and $c_2$) that constitute a one-hot vector. The remaining 98 $z$-variables are uniformly distributed on the interval  $(-1, 1)$. The Generator learns to take as the input the 2 code variables and the 98 latent variables and output 16384 samples that constitute just over one second of audio file sampled at 16 kHz through five convolutional layers. The Discriminator network takes real and generated data (again in the form of 16384 samples that constitute just over one second of audio file) and learns to estimate the Wasserstein distance between generated and real data (according to the proposal in \citealt{arjovsky17}) through five convolutional layers. In the majority of InfoGAN proposals, the Discriminator and the Q-network share convolutions.  \citet{begusCiw} introduces a separate Q-network (also in \citealt{signnet}).\footnote{For all details about the architecture, see \citet{begusCiw}.}

The Q-network is in its structure identical to the Discriminator network, but the final layer is fully connected to nodes that correspond to the number of categorical variables \citep{begusCiw}. In the ciwGAN architecture, the Q-network is trained on estimating the latent code variables with a softmax function \citep{begusCiw}.  In other words, the Q-network takes the Generator's outputs (waveforms) and estimates the Generator's latent code variables $c_1$ and $c_2$.  Weights of both the Generator network and the Q-network are updated according to the Q-network's loss function: to minimize the distance between the actual one-hot vector ($c_1$ and $c_2$) used by the Generator and the one-hot vector estimated with a softmax in the Q-network's final layer using cross-entropy. This forces the Generator to output informative data.

 The advantage of the ciwGAN architecture is that the network not only learns to output innovative data that resemble speech in the input, but also provides meaningful representations about data in an unsupervised manner.  For example, as will be argued in Section \ref{CiwGAN}, the ciwGAN network encodes reduplication as a meaningful category: it learns to assign a unique code for bare and reduplicated items. This encoding emerges in an unsupervised fashion from the requirement that the Generator output data such that unique information is retrievable from its acoustic outputs. Given the structure of the training data, the Generator is most informative if it encodes presence of reduplication in the code variables.

To replicate the results and to test learning of an identity-based pattern without binarization and without the requirement on the Generator to output informative data, we run an independent experiment on a bare WaveGAN \citep{donahue19} architecture using the same training data. The difference between the two architectures is that the bare GAN architecture does not involve a Q-network and the latent space only includes latent variables uniformly distributed on the interval $(-1, 1)$.  

\citet{begus19} and \citet{begusCiw} also propose a technique for latent space interpretability in GANs: manipulating individual variables to values well beyond the training range can reveal underlying representations of different parts of the latent space. We use this technique throughout the paper to evaluate learning of reduplication.

\section{\label{procedures} Reduplication in training data}

The training data was constructed to test a simple reduplication pattern, common in human languages: partial CV reduplication found in languages such as Paamese, Roviana, Tawala, among others \citep{inkelas05}.  Base items are of the shape C$_1$V$_2$C$_3$V$_4$ (C = consonant; V = vowel), e.g.~/tala/. Reduplicated forms are of the shape C$_1$V$_2$C$_1$V$_2$C$_3$V$_4$, where the first syllable (C$_1$V$_2$) is repeated. The items were constructed so that C$_1$ contains a voiceless stop /p, t, k/, a voiced stop /b, d, g/, a labiodental voiced fricative /v/, and nasals  /m, n/.  The vowels V$_2$ and V$_4$ consist of /\textipa{A} (\textipa{@}), i, u/. C$_3$ consists of /l, \textipa{\*r}, j/. All permutations of these elements were created. The stress was always placed on V$_2$ in the base forms and on the same syllable in reduplicated forms ([\textipa{"p\super hAl@}] $\sim$ [\textipa{p@"p\super hAl@}]). Because the reader of the training data was a speaker of American English, the training data is phonetically even more complex. The major phonetic effects in the training data include (i) reduction of the vowel in the unstressed reduplicated forms and in the final syllable (e.g.~from [\textipa{A}] to [\textipa{2/@}]) and (ii) deaspiration of voiceless stops in the unstressed reduplication syllable (e.g.~from \textipa{[p\super h]} to \textipa{[p]}). The training data includes two unique repetitions of each item and two repetitions of the corresponding reduplicated forms. Table \ref{illustration} illustrates the training data.

The training data also includes base forms C$_1$V$_2$C$_3$V$_4$  with the initial consonant C$_1$ being a fricative [s]. These items, however, always appear unreduplicated in the training data --- the purpose of [s]-initial item is to test how the network extends the reduplicative pattern to novel unobserved data.  All 27 permutations of sV$_2$C$_3$V$_4$ were included. To increase representation of [s]-initial words, four or five repetitions of each unique [s]-initial base were used in training.\footnote{Items [\textipa{"sala}], [\textipa{"suru}], and [\textipa{"suju}] each miss one repetition (four altogether).} Altogether 132 repetitions of the 27 unique unreduplicated words with an initial [s] were used in training. 

Sibilant fricative [s] was chosen as C$_1$ for testing learning of reduplication because its frication noise is acoustically prominent and sufficiently different from C$_1$s in the training data both acoustically and phonologically. This satisfies the requirement that a model learns to generalize to novel segments and feature values  \citep{berent13,prickett18}.\footnote{For an ``across the board'' generalization, \citet{berent13} requires that generalization occur to  segments fully absent from the inventory. It is challenging to elicit reduplication of segments that are fully absent from the training data in the proposed models. Even in human subject experiments testing the ``across the board'' generalization, subjects need to be exposed to the novel segment at least as a prompt. In our case, the novel segment needs to be part of the training data, but only in unreduplicated forms.} In phonological terms, the model is tested on a novel feature (sibilant fricative or [$\pm$strident]; \citealt{hayes09}) --- the training data did not consist of any bare or reduplicated forms  with other sibilant fricatives. To make the learning even more complex, voiceless fricatives ([f, \textipa{T}, \textipa{S}]) are altogether absent from the training data. All voiced fricatives except for [v] are absent too. Spectral properties of the voiced non-sibilant fricative [v] in the training data (and in Standardized American English in general) are so substantially different from a voiceless sibilant fricative [s] that we kept them in the training data. We excluded all items with initial sequences /ti/, /tu/, and /ki/ from the training data, because acoustic properties of these sequences, especially frication of the aspiration of /t/ and /k/, are similar to those of frication noise in /s/. Altogether 996 unique sliced items used in training were recorded in a sound attenuated booth  by a female speaker of American English with a MixPre 6 (SoundDevices) preamp/recorder and the AKG C544L head-mounted microphone.

\begin{table}
\centering
\begin{tabular}{lll}
\hline\hline
\multirow{2}{*}{voiceless C$_1$}&C$_1$V$_2$C$_3$V$_4$&\textipa{"p\super hAli}\\
&C$_1$V$_2$C$_1$V$_2$C$_3$V$_4$&\textipa{\phantom{"}p2"p\super hAli}\\
\hline
\multirow{2}{*}{voiced C$_1$}&C$_1$V$_2$C$_3$V$_4$&\textipa{"bAli}\\
&C$_1$V$_2$C$_1$V$_2$C$_3$V$_4$&\textipa{\phantom{"}b2"bAli}\\
\hline
\multirow{2}{*}{C$_1$ = [m, n, v] }&C$_1$V$_2$C$_3$V$_4$&\textipa{"mAli}\\
&C$_1$V$_2$C$_1$V$_2$C$_3$V$_4$&\textipa{\phantom{"}m2"mAli}\\

\hline
\multirow{2}{*}{C$_1$ = [s]}&C$_1$V$_2$C$_3$V$_4$&\textipa{"sAli}\\
&C$_1$V$_2$C$_1$V$_2$C$_3$V$_4$&---\\
\hline\hline
\end{tabular}
\caption{A schematic illustration of the training data in the International Phonetic Alphabet. 
}
\label{illustration}
\end{table}

\section{CiwGAN \citep{begusCiw}}
\label{CiwGAN}

The Generator features two latent code variables, $c_1$ and $c_2$ and 98 uniformly distributed variables $z$ (Figure \ref{ciwRedgantikz}). In the training phase, the two code variables ($c_1$ and $c_2$) compose the one-hot vector with two levels: [0,~1] and [1,~0]. This means that the network can encode two categories in its latent space structure that correspond to some meaningful feature about the data. The Q-network forces the Generator to encode information in its latent space. In other words, the loss function of the Q-network forces the Generator to output data such that the Q-network is effective in retrieving the latent code $c_1$ and $c_2$ from the Generator's acoustic outputs only. Nothing in the training data pairs base and reduplicated forms. There is no overt connection between the bases and their reduplicated correspondents. Yet, the structure of the data is such that given two categories, the most informative way for the Generator to encode unique information in its acoustic outputs is to associate one unique code with base forms and another with reduplicated forms. The Generator would thus have a meaningful unique representation of reduplication that arises in an unsupervised manner exclusively from the requirement on the Generator to output informative data.

To test whether the Generator encodes reduplication in latent codes, we train the network for 15,920 steps (or approximately 5,114 epochs) with the data described in Section \ref{procedures}. The choice of the number of steps is based on two objectives; first, the output data should approximate speech to the degree that allows acoustic analysis. Second, the Generator network should not be trained to the degree that it replicates data completely. As such, overfitting rarely occurs in the GAN architecture \citep{adlam19,donahue19}.  The best evidence against overfitting in the ciwGAN architecture comes from the fact that the Generator outputs data that violate training distributions substantially (see Section \ref{unobserved} below) \citep{begusCiw,begusLocal}.  Despite these guidelines, the choice of number of steps  is somewhat arbitrary (for discussion, see \citealt{begus19}).

We generate 100 outputs for each latent code [0,~1] and [1,~0] (200 total) and  annotate them for presence or absence of reduplication. All annotations here and in other sections are performed by the author in Praat \citep{boersma15}. Distinguishing unreduplicated from reduplicated is very salient; for less salient annotations, we provide waveforms and spectrograms (e.g.~Figures \ref{sisiyugenhuman} and \ref{figures}).\footnote{The code is available at \url{https://github.com/gbegus/fiwGAN-ciwGAN}. The generated data and checkpoints are available at \url{https://doi.org/10.17605/osf.io/zbjcp}.} 

There is a significant correlation between the two levels of latent code and presence of reduplication. Counts are given in Table \ref{rcount}. When the code is set to [1,~0], 78\% of the generated outputs are base forms; when set to [0,~1], 60\% of outputs are reduplicated  (odds ratio = 5.27, $p<0.0001$, Fisher Exact Test). When the latent codes are set to [0,~5] and [5,~0], we get  a near categorical distribution of bare and reduplicated forms. For  [5,~0], the Generator outputs an unreduplicated bare form in 98\% samples. For [0,~5], it outputs a reduplicated form in 87\% outputs (odd ratio = 308.3, $p<0.0001$, Fisher Exact Test). These outcomes suggest that the Generator encodes reduplication in its latent codes and again confirm that manipulating latent variables well beyond training range reveals the underlying  learning representations in deep convolutional networks (as proposed in \citealt{begus19,begusCiw}).

\begin{table}\centering
\begin{tabular}{rccc}
\hline\hline
\textbf{Code}&\textbf{Bare}&\textbf{Redup.}&\textbf{\% Redup.}\\\hline
\textbf{[1, 0]}& 78 &22& 22\%\\
\textbf{[0, 1]} &40&60&60\% \\\hline
\textbf{[5, 0]} &98&2&2\% \\
\textbf{[0, 5]} &13&87&87\% \\

\hline\hline
\end{tabular}
\caption{Counts of bare and reduplicated (redup.) outputs when the latent codes $c_1$ and $c_2$ are set to [1, 0], [0, 1], [5, 0], and [0, 5].}
\label{rcount}
\end{table}

\subsection{Interpolation}
\label{interpolation}

That the Generator uses latent codes to encode reduplication is further suggested by another generative test performed on interpolated values of the latent code. To test how exactly the relationship between the latent codes ($c_1$ and $c_2$) works, we created sets of generated outputs based on interpolated values of the code  $c_1$ and $c_2$. We manipulate $c_1$ and $c_2$ from the value 1.5 towards 0 in increments of 0.125. For example, we start with [1.5,~0] and interpolated first to [0,~0] (e.g.~[1.375,~0], [1.25,~0], etc.). From [0,~0] we further interpolate in increments of 0.125 to [0,~1.5] (e.g.~[0,~0.125], [0,~0.25]). All other variables in the latent space are kept constant across all interpolated values. Each such set thus contains 25 generated samples. We generate 100 such sets (altogether 2500 outputs)  and analyze each output. Out of the 100 sets, the output was either bare or reduplicated throughout the interpolated values and did not change in 55 sets. As suggested by Section \ref{CiwGAN} and Table \ref{rcount}, the number of bare and reduplicated forms for each level rises to near categorical values as the variables approach values of 5.  

In the 45/100 sets, the output changes from the base form to a reduplicated form at some point as the codes are interpolated. If the network only learned to randomly associate base and  reduplicated forms with each endpoint of the latent code, we would expect base forms to be unrelated to reduplicated forms. For example, a base form [\textipa{"k\super hulu}] could turn into reduplicated [\textipa{d@"dAl@}]. An acoustic analysis of the generated sets, however, suggests that the latent code directly corresponds to reduplication. In approximately 25 out of 45  sets (55.6\%) of generated outputs that undergo the change from base to a reduplicated form (or 25\% of the total sets), the base form is identical to the reduplicated form with the only major difference between the two being the presence of reduplication (waveforms and spectrograms of the 25 outputs are in Figure \ref{figures}). This proportion would likely be even higher with a higher interpolation resolution (higher than 0.125) and because we do not count cases in which major changes of sounds occur besides the addition of the reduplication syllable (for example, if [\textipa{"nA\*ri}] changes to [\textipa{nU"nu\*ri}], we count the output as unsuccessful). In the remaining 20 outputs, several outputs undergo changes, where several segments or their features are kept constant, but the degree to which they differ can vary (e.g.~[\textipa{"p\super hil@}] $\sim$ [\textipa{p@"p\super hi\*ri}],  [\textipa{"t\super hiju}] $\sim$ [\textipa{d@"dAji}],  [\textipa{"nA\*r@}] $\sim$ [\textipa{d@"dA\*ri}], or [\textipa{"p\super hi\*r@}] $\sim$ [\textipa{t@"t\super hAli}]). 

Under the null hypothesis, if the Generator learns to pair the base and reduplicated forms randomly, each base form could be associated with any of the unique 243 reduplicated forms at the probability of 1/243 (0.004). Even if we assume very conservatively that each base form could be associated with only each subgroup of reduplicated consonant (C$_1$; e.g.~voiceless stops, voiced stops, [m], [n], [v]) disregarding the vowel  and disregarding changes in the base,  the probability of both forms being identical would still be at only 0.2 (for each of the five subgroups). In both cases, the ratio of identical base-reduplication pairs, while not categorical, is highly significant ($\text{CI} = [0.4,~0.7], p<0.0001$ for both cases according to Exact Binomial Test). 

\begin{figure}
\centering
\includegraphics[width=.25\textwidth]{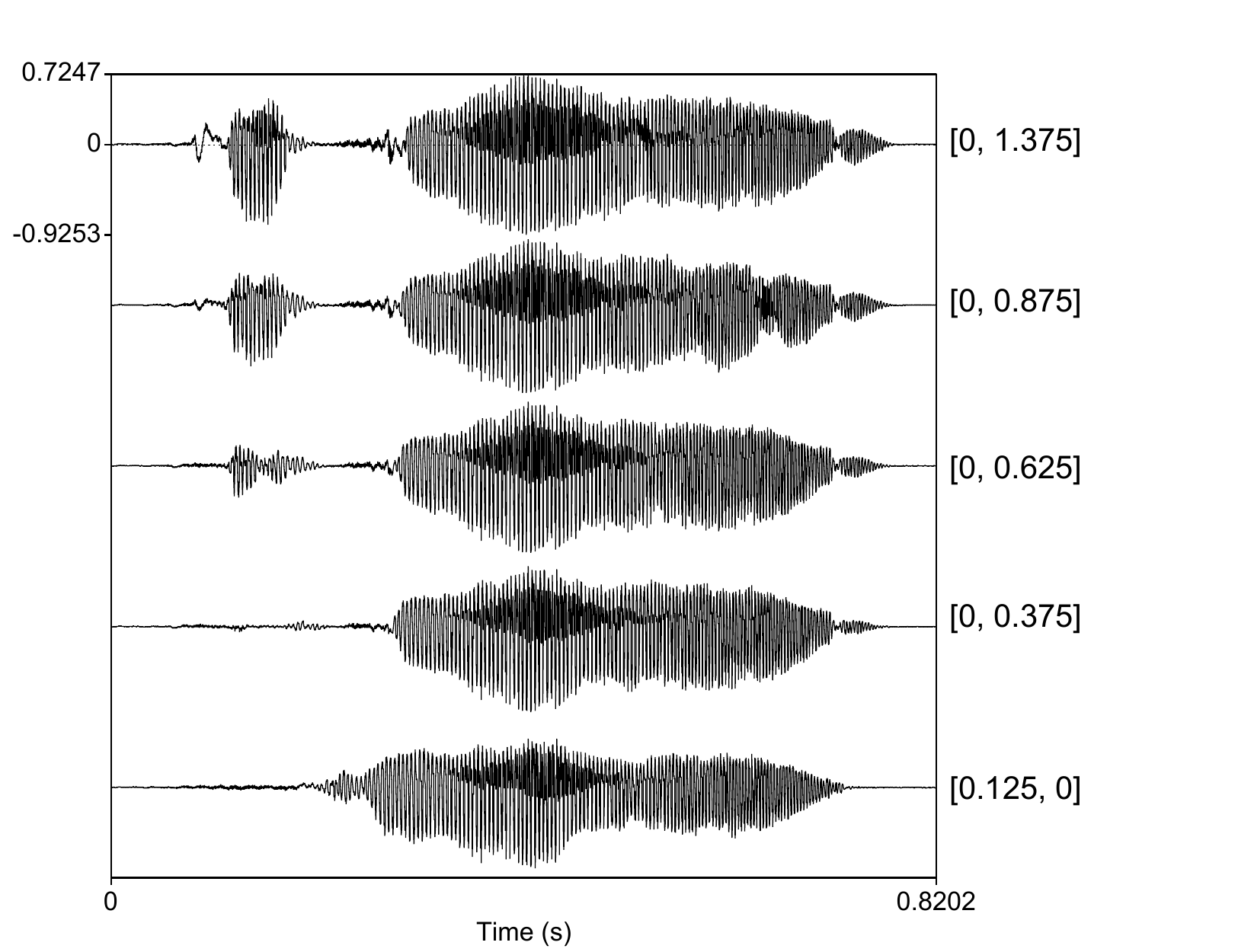}\includegraphics[width=.25\textwidth]{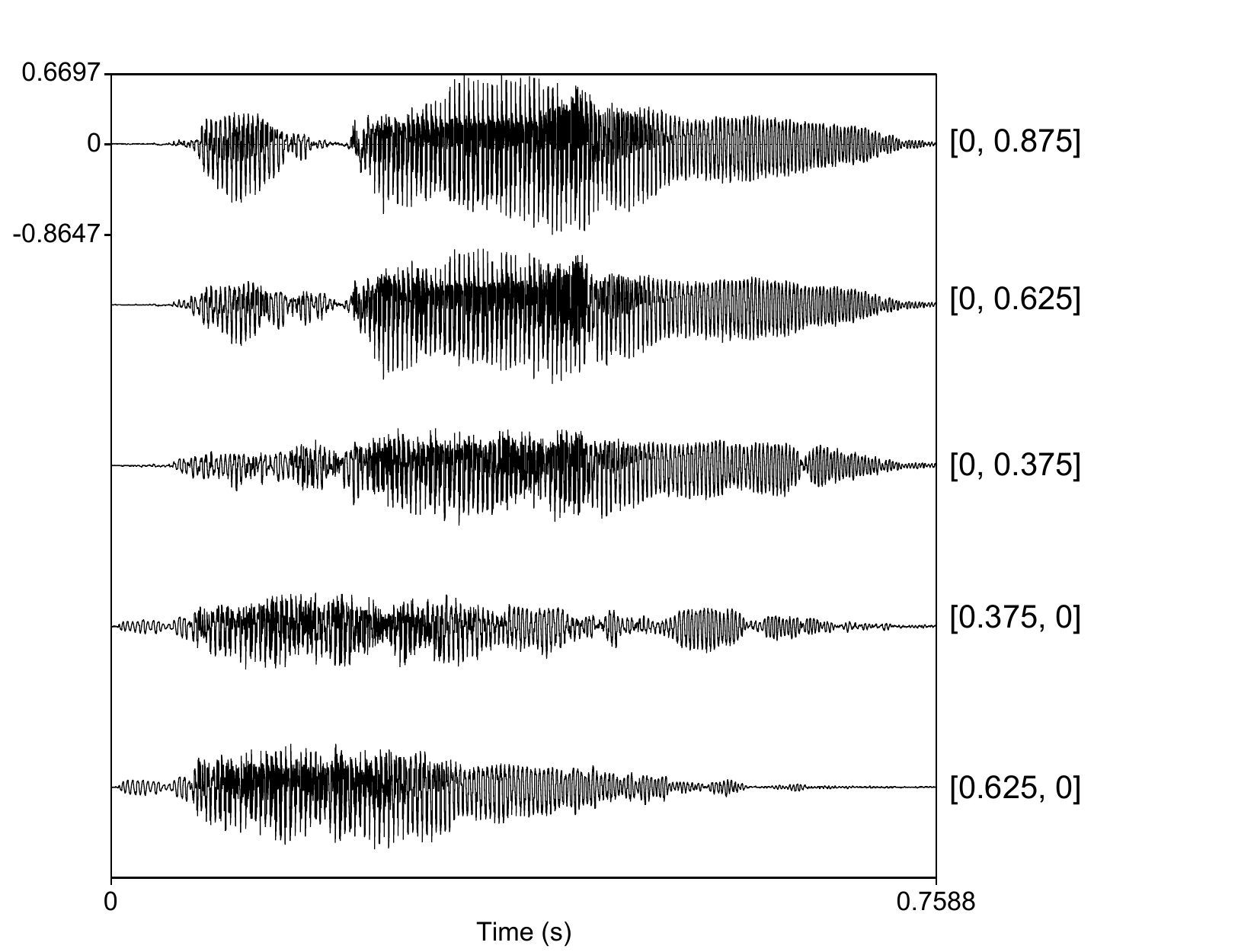}
\caption{\label{pipiruS241a} Waveforms showing how interpolation of latent codes $c_1$ and $c_2$ has a direct effect on presence of reduplicattion: as the values are interpolated from [1.5, 0] to [0, 1.5],  the reduplication gradually appears/disappears from the output. Waveforms on the left represent reduplication of [\textipa{"p\super hi\*ru}] to [\textipa{p@"p\super hi\*ru}]; waveforms on the right represent reduplication of [\textipa{"dAji}]  to [\textipa{d@"dAji}].  
}
\end{figure}

Figure \ref{pipiruS241a} illustrates how, keeping the latent space constant except for the manipulation of the latent code with which the Generator represents reduplication, the generated outputs gradually transition from the base forms [\textipa{"p\super hi\*ru}] and  [\textipa{"dAji}] to the reduplicated forms [\textipa{p@"p\super hi\*ru}] and  [\textipa{d@"dAji}].\footnote{The exact vowel quality estimation in the generated outputs is challenging, especially in short vocalic elements of reduced vowels in the reduplicative syllables. For this reason, we default transcriptions to a [\textipa{@}].} Other major properties of the output are unchanged. 

This interpolative generative test again suggests that the network learns reduplication and encodes the process in the latent codes. By interpolating the codes we can actively force reduplication in the output with no other substantial changes in the majority of cases.

\subsection{Reduplication of unobserved data}
\label{unobserved}

To test whether the ciwGAN network learns to generalize the reduplicative pattern on unobserved data, we use latent space manipulation to force reduplication at the same time as presence of [s] in the output. Items with a [s] as the initial consonants (e.g.~[\textipa{"siju}]) appear only in bare forms in the training data. In Sections \ref{CiwGAN} and \ref{interpolation}, we established that the network uses the latent code ($c_1$ and $c_2$) to represent reduplication. Following \citet{begus19} and \citet{begusCiw}, we can force any phonetic property in the output by manipulating the latent variables well beyond the training range. Reduplication is forced by setting the latent code to values  higher than [0, 1]. We can simultaneously force [s] in the output to test the network's performance on reduplication in unseen data. 

\begin{figure}
\centering
\includegraphics[width=.47\textwidth]{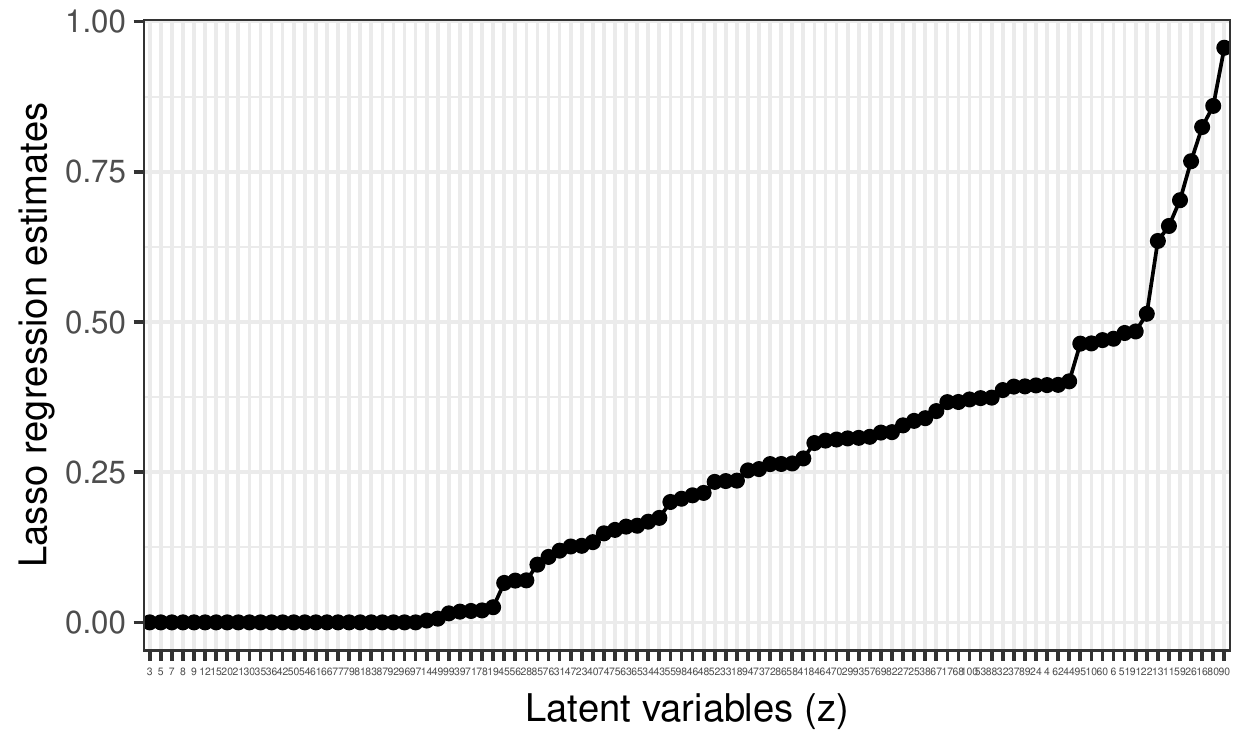}
\caption{\label{redGANciwGANs} Absolute Lasso regression estimates (sorted from highest on the right-hand side) for  a ciwGAN model identifying presence of [s] after 1000 transcribed outputs, 500 for each latent code (with the same latent variable structure of the remaining 98 variables across the two conditions). Variable $z_{90}$ is identified as the variable corresponding to presence of [s] (the variable with the highest regression estimates). }
\end{figure}

To identify latent variables with which the Generator encodes the sound [s] in the output, we generate 1000 samples with randomly sampled latent variables, but with the latent code variables ($c_1$ and $c_2$) set at [0, 1] and [1, 0] (500 samples each with the same latent variable structure of the remaining 98 variables across the two conditions). We annotate outputs for presence of [s] for the two sets and fit the data to a Lasso logistic regression model in the \emph{glmnet} package \citep{glmnet}. Presence of [s] is the dependent variable coded as a success; the independent variables are the 98 latent variables uniformly distributed on the interval $(-1, 1)$ (for the technique, see \citealt{begus19}). Lambda is computed with 10-fold cross validation. Estimates of the Lasso regression model (Figure \ref{redGANciwGANs}) suggest that $z_{90}$ with the highest regression estimates is one of the variables with which the Generator encodes presence of [s] in the output. For a generative test providing evidence that Lasso regression estimates correlate with network's internal representations, see \citet{begus19}.

\begin{figure*}
\centering
\includegraphics[width=.35\textwidth]{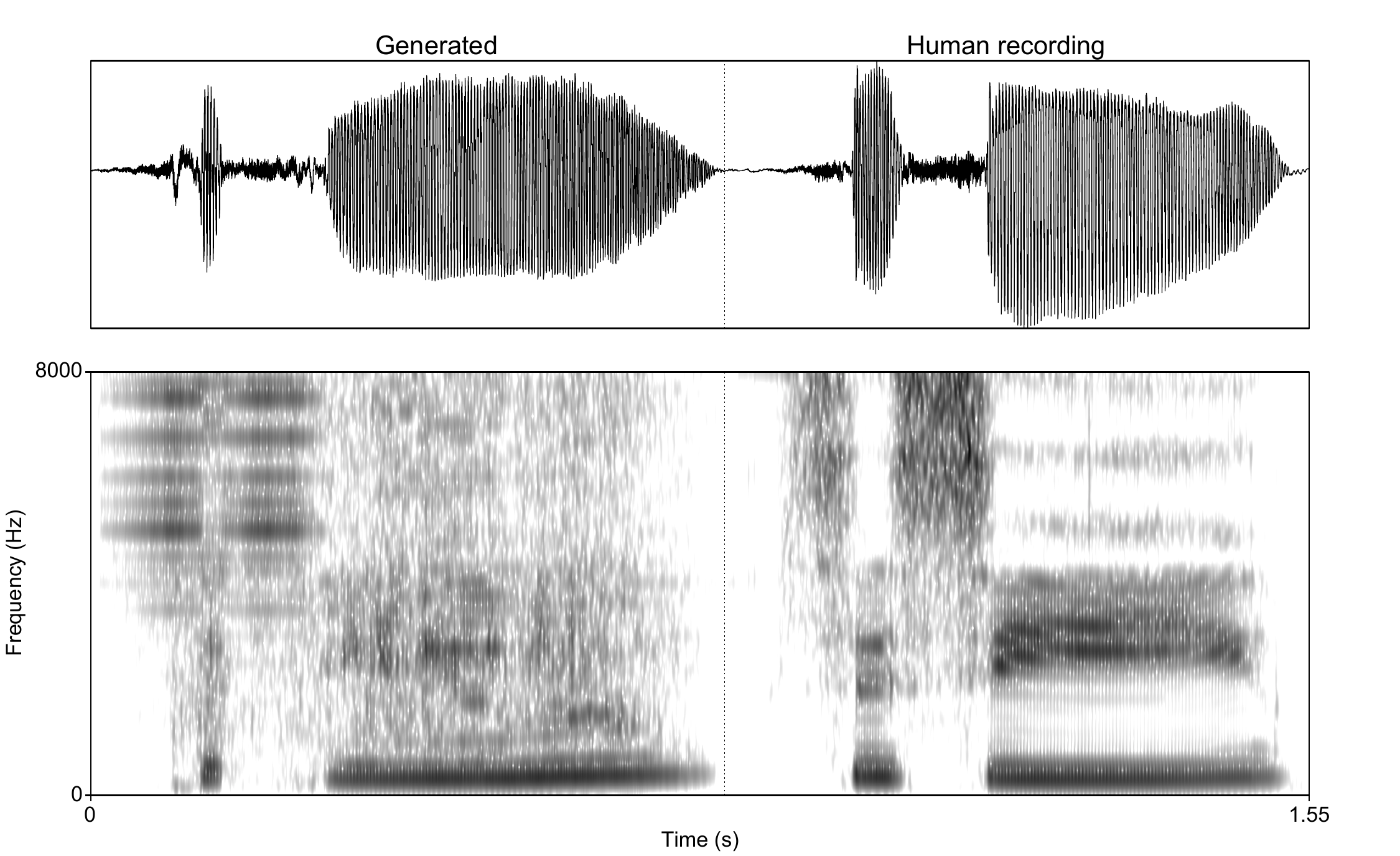}
\includegraphics[width=.35\textwidth]{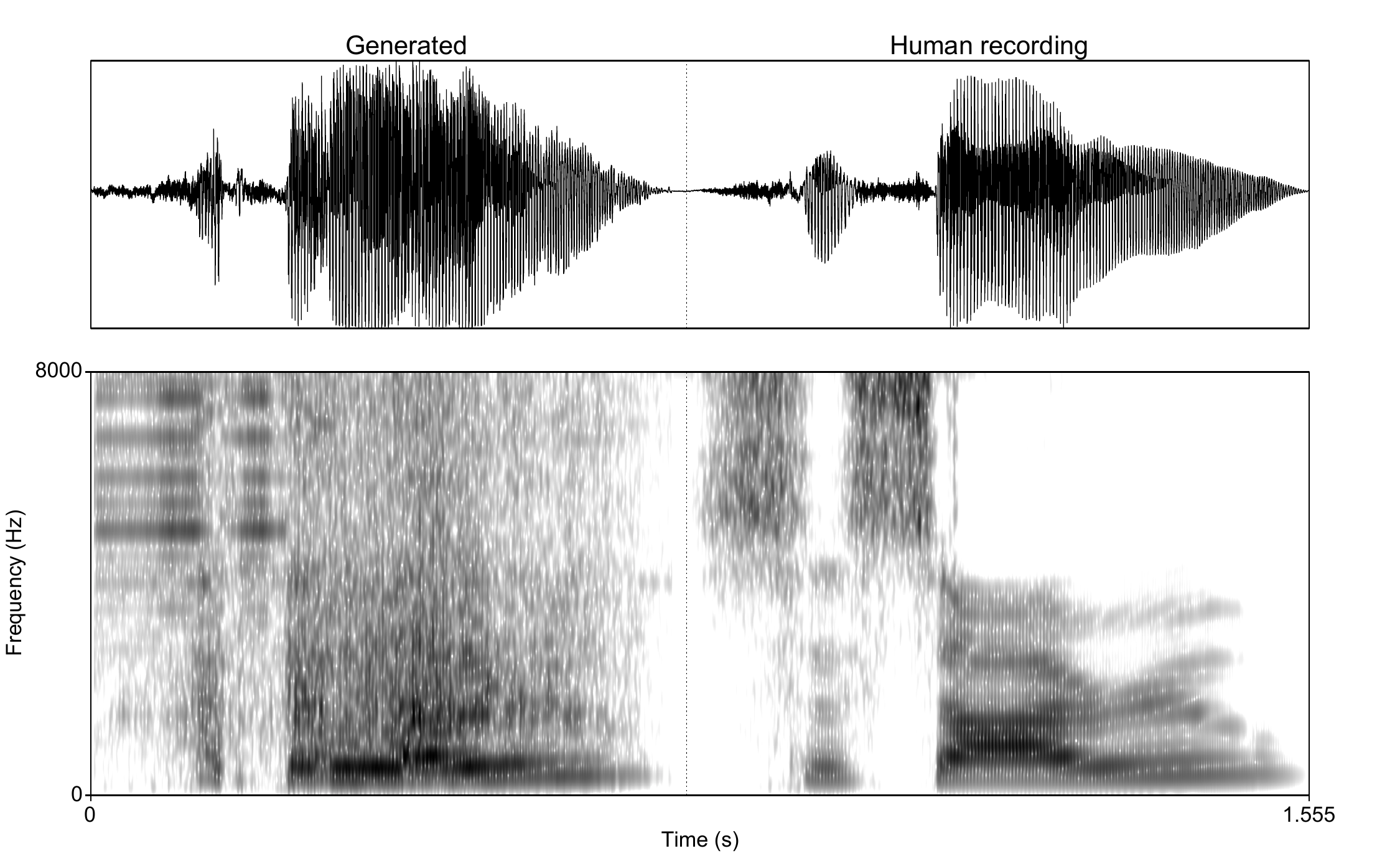}
\includegraphics[width=.35\textwidth]{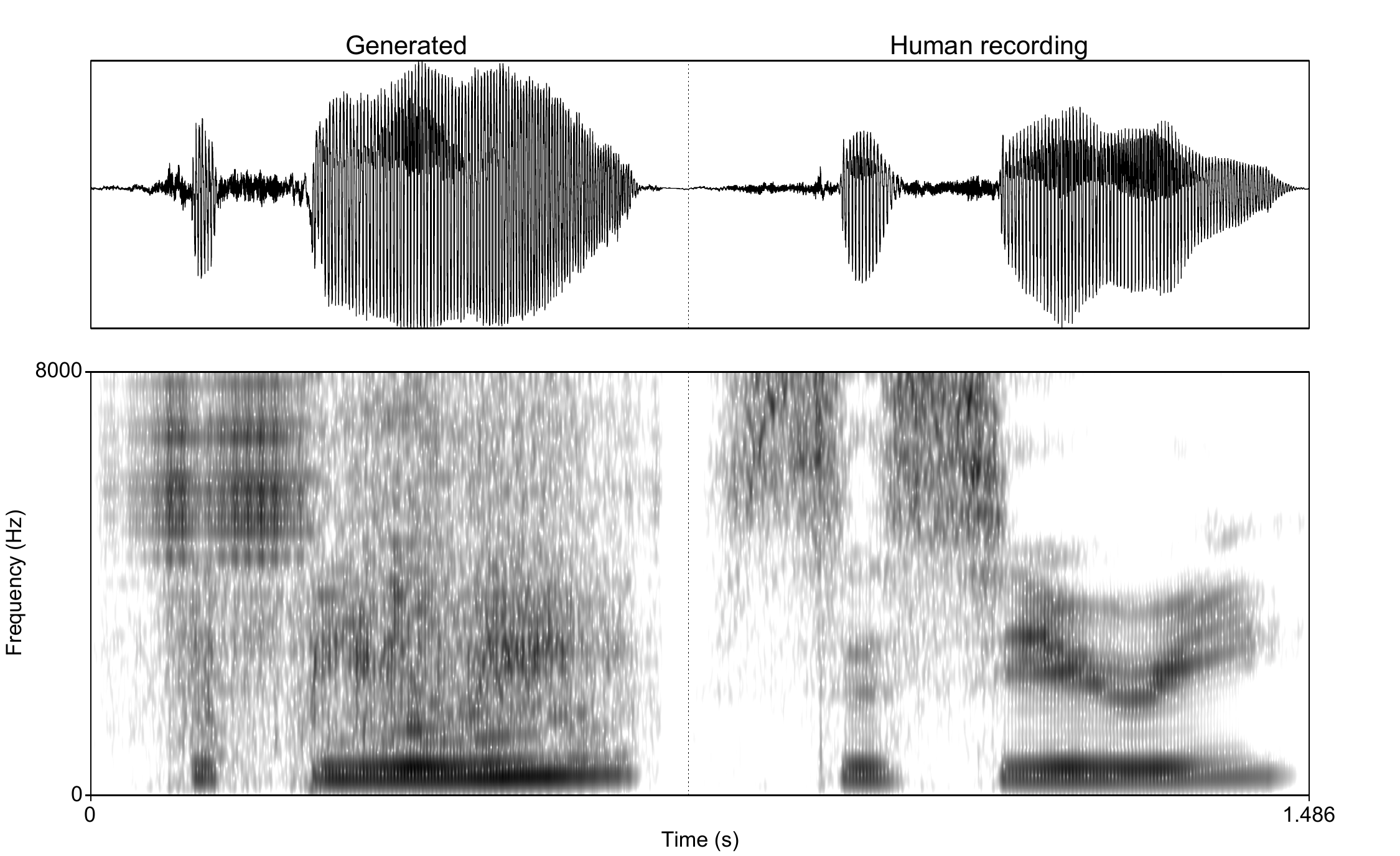}
\caption{\label{sisiyugenhuman} Waveforms and spectrograms (0--8000 Hz) of reduplicated forms containing an [s] which were absent from the training data. The generated forms on the left are paired with recordings of a female speaker reading reduplicated forms that were absent from the training data. (left)   When the latent code is set to [0, 3] and $z_{90}$ to 4, the network outputs a reduplicated [\textipa{s@"siji}]. (right)  When the latent code is set to [0, 7.5] and $z_4$ to 9.5, we get [\textipa{s@"sA\*ru}]. 
(bottom)   In the bare GAN architecture, when  $z_{5}$ (forcing reduplication) is set to $-9.25$ and $z_{17}$ (forcing [s] in the output)   to $-9.0$, the Generator outputs a reduplicated [\textipa{s@"si\*ri}]. 
 }
\end{figure*}

We can thus set $z_{90}$ to marginal levels well beyond the training range and the latent code ($c_1$, $c_2$) to levels well beyond [0, 1] in order to force reduplication and [s] in the output simultaneously. For example, when the latent code is set to [0, 3] (which forces reduplication in the output) and $z_{90}$ to 4 (forcing [s] in the output), the network outputs a reduplicated [\textipa{s@"siji}] (among other outputs) even though items containing an [s] are never reduplicated in the training data. When the code is set to even higher number, [0, 7.25], and $z_{90}$ to 7, the network outputs [\textipa{s@"siru}] in a different output. The spectrograms in Figure \ref{sisiyugenhuman} show a clear period of frication noise characteristic of a sibilant fricative [s], interrupted by a reduplicative vowel and followed by a repeated period of frication noise characteristic of [s].

In fact, at the values [0, 7.25], and $z_{90}= 7$, the network generates approximately 33 (out of 100 tested or 33\%) outputs that can be reliably analyzed as reduplicated forms with initial sV- reduplication unseen in the training data. The other 67 outputs are reduplicated forms containing other C$_1$s or unreduplicated [s]-forms. No outputs were observed in which C$_1$ of the reduplication syllable and C$_1$ of the base would be substantially different. While all the cases when  $z_{90}$ is manipulated involve a front vowel [i] in the base item, we can also elicit reduplication for other vowels. For example, we identify variable $z_4$ as corresponding to an [s] and a low vowel \textipa{[A]} in the output (with the same technique as described for $z_{90}$ above but with presence of [s\textipa{A}] as the dependent variable in the Lasso regression model). By manipulating $z_4$ to 9.5 (forcing [\textipa{sA}] in the output) and setting the latent codes to [0, 7.5], we get [\textipa{s@"sA\*ru}]  in the output (Figure \ref{sisiyugenhuman}).

For comparison, the same L1 speaker of English who read the words in the training data read the reduplicated  [\textipa{s@"siji}] and [\textipa{s@"sA\*ru}]  which were not included in the training data. Figure \ref{sisiyugenhuman} parallels the generated reduplicated forms based on unobserved data (which were elicited by forcing [s] and reduplication in the output) and the recording of the same reduplicated form read by a human speaker. The spectrograms show clear acoustic parallels between the Generated outputs and the recording read by a human speaker (who read the words prior to computational experiments and did not hear or analyze the generated outputs).

\section{Replication: Bare WaveGAN \citep{donahue19}}
\label{reproduction}

To test whether the learning of reduplicative patterns in GANs is a robust or idiosyncratic property of the model presented in Section \ref{CiwGAN}, we conduct a replication experiment. We introduce two crucial differences in the replication experiment: we train the Generator without the requirement to produce informative data and without binary latent codes. We use the model in \citet{donahue19} which features a ``bare'' GAN architecture for audio data: only the Generator and Discriminator networks without the Q-network. This architecture has the potential to inform us how GANs represent reduplicative patterns without an explicit requirement to learn informative data, i.e.~without an explicit requirement to encode some salient feature of the training data in the latent space. The data used for training is the same as in the experiment  in Section \ref{procedures}. We train the network for 15,930 steps or approximately 5,118 epochs, which is almost identical to the number of steps/epochs in the ciwGAN experiment (Section \ref{CiwGAN}).

\subsection{\label{variables} Identifying variables}

\begin{figure*}
\centering
\includegraphics[width=.7\textwidth]{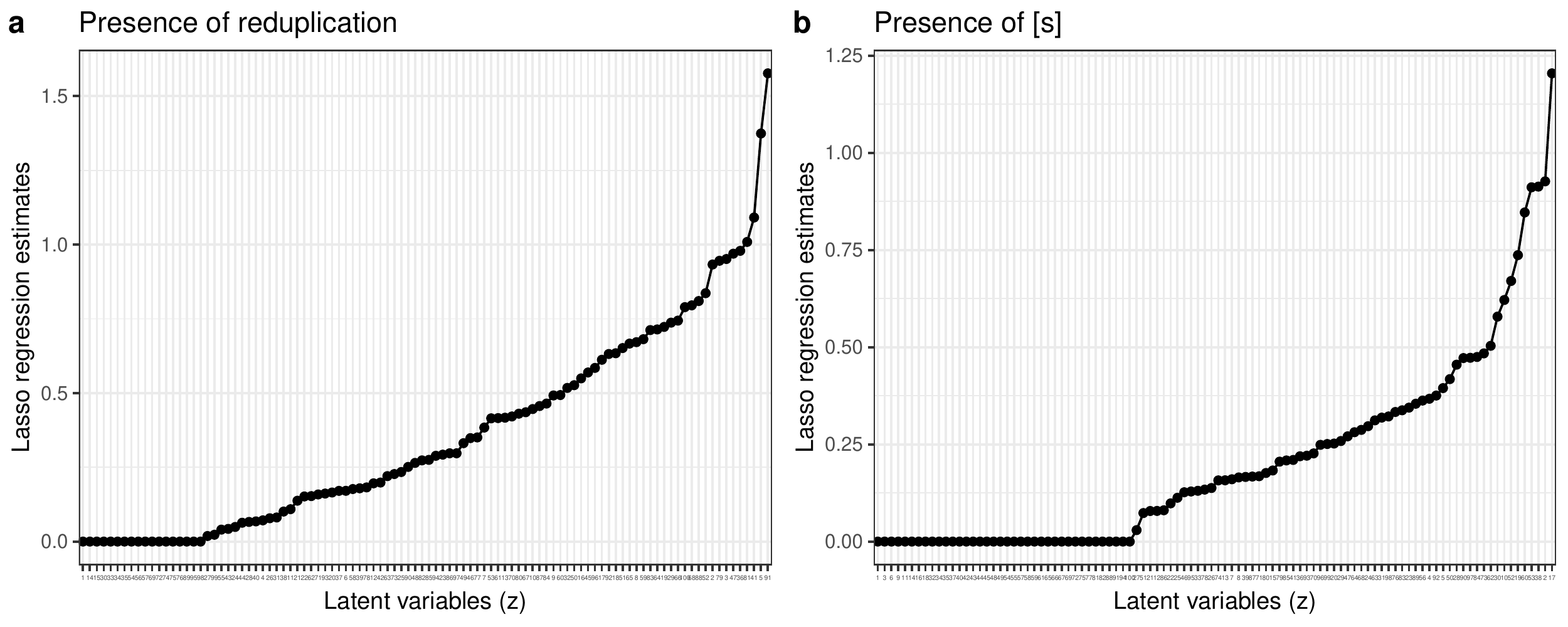}
\caption{\label{redGANredinSa} Absolute Lasso regression estimates (sorted from highest on the right-hand side) for two models identifying (a) presence of reduplication and (b) presence of [s] in the generated outputs of the bare GAN model (Section \ref{reproduction}). 
}
\end{figure*}

Testing the learning of reduplication in the bare GAN architecture requires that we force reduplication and presence of [s] in the output simultaneously. To identify which latent variables correspond to the two properties, we use the same technique  as described in Section \ref{CiwGAN}. We generate and annotate 500 outputs of the Generator network with randomly sampled latent variables. We annotate the presence of [s] and the presence of reduplication. The annotations are fit to a Lasso logistic regression (as in Section \ref{unobserved}): presence of reduplication or [s] are the dependent variables and each of the 100 latent $z$-variables are the independent predictors. Lambda values were computed with 10-fold cross validation. Regression estimates are given in Figure \ref{redGANredinSa}. 

The plots illustrate a steep drop in regression estimates between the few latent variables with the highest estimates and the rest of the latent space. In fact, in both models, one or two variables per model emerge with substantially higher regression estimates: $z_{91}$ and $z_{5}$ when the dependent variable is \textsc{presence of reduplication} and $z_{17}$ when the dependent variable is \textsc{presence of} [s] in the output. We can assume the Generator network uses these two variables to encode presence of reduplication and [s], respectively. 

It has been argued in \citet{begus19} that GANs learn to encode phonetic and phonological representations with a subset of latent variables. The discretized representation of continuous phonetic properties in the latent space appears even more radical in the present case. For example, in \citet{begus19}, presence of [s] as a sound in the output is represented by at least seven latent variables, each of which likely controls different spectral properties of the frication noise. In the present experiment, the Generator appears to learn to encode presence of [s] with a single latent variable, as is suggested by a steep drop of regression estimates after the first variables with the highest estimates. For a generative test showing that regression estimates correlate to actual rates of a given property in generated data, see \citet{begus19}. Such near-categorical cutoff is likely a consequence of the training data in the present case being considerably less variable compared to TIMIT (used for training in \citealt{begus19}). The network also represents an identity-based process, reduplication, with only two latent variables and features a substantial drop in regression estimates after these two variables.  This discretized representation thus emerges even without the requirement of the Generator to output informative data.

In the replication experiment too, the Generator network outputs reduplicated forms for unobserved data when both reduplication and [s] are forced in the output via latent space manipulation, but significantly less so than in the ciwGAN architecture. When $z_{91}$ (forcing reduplication) and $z_{17}$ (forcing [s] in the output) are set to value $-8.5$, a higher level compared to the generated samples in the ciwGAN architecture (7 and 7.25), the network outputs only one reduplicated form with [s]-reduplication out of 100 generated outputs. By comparison, the proportion of the [s]-reduplication in the ciwGAN architecture is 33/100 -- a significantly higher ratio ($\text{OR} = 48.1, p<0.0001$; Fisher Exact Test). When  $z_{5}$ (forcing reduplication) is set to $-9.25$ and $z_{17}$ (forcing [s] in the output)   to $-9.0$, the proportion of reduplicated [s]-items is slightly higher (4/100), but still significantly lower than in the ciwGAN architecture  ($\text{OR} = 11.7, p<0.0001$; Fisher Exact Test). Despite these lower proportions of reduplicated [s] in the output, the bare GAN network nevertheless extends reduplication on novel unobserved data. Figure \ref{sisiyugenhuman} illustrates an example of a reduplicated [s]-item from the Generator network trained in the bare GAN architecture: [\textipa{s@"si\*ri}]. The spectrogram reveals a clear period of frication noise characteristic of an [s], followed by a reduplicative vowel period, followed by another period of frication.

\section{Discussion}

We perform four generative tests to model learning of reduplication in deep convolutional networks: (i) a test of proportion of outputs when latent codes are manipulated to marginal values, (ii) a test of interpolating latent variables, (iii) a test  of reduplication on unobserved data in the ciwGAN architecture, and (iv) a replication test  of reduplication on unobserved data in the bare WaveGAN architecture. All four tests suggest that deep convolutional networks learn a simple identity-based pattern in speech called reduplication, i.e.~a process that copies some phonological material to express new meaning.   
The ciwGAN network learns to encode a meaningful representation --- presence of reduplication into its latent codes. There is a near one-to-one correspondence between the two latent codes $c_1$ and $c_2$ and reduplication. By interpolating latent codes, we cause the bare form to gradually turn into a reduplicated form with no other major changes in the output in the majority of cases. These results are close to what would be considered appearance of symbolic computation or algebraic rules. 

Additional evidence that an approximation of symbolic computation emerges comes from the bare GAN experiment: there is a substantial drop in regression estimates after the first one or two latent variables with highest regression estimates, suggesting that even without the requirement to produce informative data, the network discretizes the continuous and highly variable phonetic feature --- presence of reduplication --- and uses a small subset of the latent space to represent this morphophonological property. 

Finally, we can force the Generator to output reduplication at nearly categorical levels. When latent codes are set to marginal levels outside of training range (e.g.~to [5,~0] or [0,~5]), the outputs are almost categorically unreduplicated or reduplicated (at 98 \% for [5,~0]). \citet{begusCiw} shows that even higher values (e.g.~15) result in performance at 100\% for a subset of variables. Not all aspects of the models in this paper are categorical (e.g.~interpolation of latent codes does not always change an  unreduplicated to a reduplicated form without other major changes). Improving performance on this particular task is left for future work. Inability to derive categorical processes has long been an argument against the connectionist approaches to language modeling. The results of this experiments add to the work suggesting that manipulating variables to extreme marginal values results in near categorical or categorical  outputs (depending on the value) of a desired property \citep{begus19,begusCiw}.

In sum, three properties of rule-like symbolic representations emerge in deep convolutional network tested here: discretized representations, the ability to generate desired property by manipulating a small number of variables, and near categoricity for a subset of representations. These symbolic-like outcomes are facilitated by two inductive biases: the binary nature of latent codes and the requirement on the Generator to output informative data (forced by the Q-network). At least a subset of these properties also emerges in the bare WaveGAN architecture that lacks these biases, but at a reduced performance. 

Encoding an identity-based pattern as a meaningful representation in the latent space emerges  in a completely unsupervised manner in the ciwGAN architecture --- only from the requirement that the Generator output informative data.  Reduplicated and unreduplicated forms are never paired in the training data. The network is fed bare and reduplicated forms randomly. This unsupervised training approximates conditions in language acquisition (for hearing learners): the human language learner needs to represent reduplication and to pair bare and reduplicated forms from raw  unlabeled acoustic data. The ciwGAN learns to group  reduplicated and unreduplicated forms and assign a unique representation to the process of reduplication. In fact,  the one-hot vector ($c_1$ and $c_2$) that the Generator learns to associate with reduplication in training can be modeled as a representation of the unique meaning/function that reduplication adds, in line with an approach to represent unique semantics with one-hot vectors   (e.g.~in \citealt{shane20}).

The paper also argues that deep convolutional networks can learn a simple identity-based pattern (copying) from raw continuous data and extend the pattern to novel unobserved data. 
While the network was not trained on reduplicated items that start with an [s], we were able to elicit reduplication in the output following a technique proposed in  \citet{begus19}. First, we identify variables that correspond to some phonetic/phonological representation such as presence of [s]. We argue that setting single variables well above training range can reveal the underlying value for each latent variable and force the desired property in the output. We can thus force both [s] and reduplication in the output simultaneously. For example, the network outputs [\textipa{s@siju}] if we force both reduplication and [s] in the output;  however, it never sees [\textipa{s@siju}] in the training data --- only  [\textipa{siju}] and other reduplicated forms, none of which included an [s]. 

Thus, these experiments again confirm that the network uses individual latent variables to represent linguistically meaningful representations \citep{begus19,begusCiw}. Setting these individual variables to values well above the training interval reveals their underlying values. By manipulating these individual variables, we can explore how the representations are learned as well as how interactions between different variables work (for example, between the representation of reduplication and presence of [s]). The results of this study suggest that the deep convolutional network is not only capable of encoding different phonetic properties in individual latent variables, but also processes as abstract as copying or reduplication. 

One of the advantages of probing learning in deep convolutional neural networks on speech data trained with GANs is that the innovative outputs violate training data in structured and highly informative ways. The innovative outputs with reduplication of [s]-initial forms such as [\textipa{s@siju}] can be directly paralleled to acoustic outputs read by L1 speaker of American English  that were absent from the training data. Acoustic analysis shows a high degree of similarity between the generated reduplicated forms and human recordings, meaning that the network learns to output novel data that are linguistically interpretable and resemble human speech processes even though they are absent from the training data. Thus, the results of the experiments have implications for cognitive models of speech acquisition. It appears that one of the processes that has long been held as a hallmark of symbolic computation in language, reduplication, can emerge in deep convolutional network without language-specific components in the model even when they are trained on raw acoustic inputs. 

The present paper tests a simple partial reduplicative pattern where only CV is copied and appears before the base item.  This is perhaps computationally the simplest reduplicative pattern. The training data are also highly controlled and recorded by a single speaker. We can use the well-understood identity-based patterns in speech with various degrees of complexity (longer reduplication, embedding into non-reduplicative patterns) to further test how  inductive biases and hyperparameter/architecture choices interact with learning in deep convolutional networks. Finally, learning biases in the ciwGAN model can be (superficially) compared to learning biases in human subjects in future work. This paper suggests that the Generator provides informative outputs even if trained on comparatively small data sets (for a similar conclusion for other processes, see \citealt{begusLocal}). This means we can use the same training data to probe learning in CNNs and in human artificial grammar learning experiments (for a methodology, see  \citealt{begusLocal}). While these comparisons are necessarily superficial at this point, they can provide insights into common learning biases between human learners and computational models.

\section*{Acknowledgements}
This work was supported by a grant to new faculty at the University of Washington. I would like to thank Ella Deaton for recording and preparing stimuli as well as anonymous reviewers and the Action Editor for useful comments on earlier versions of this paper.

\bibliography{begusGANbib.bib}
\bibliographystyle{acl_natbib}


\onecolumn
\appendix

\section{Appendix} 

\begin{figure}[H]
\centering
\includegraphics[width=.19\textwidth]{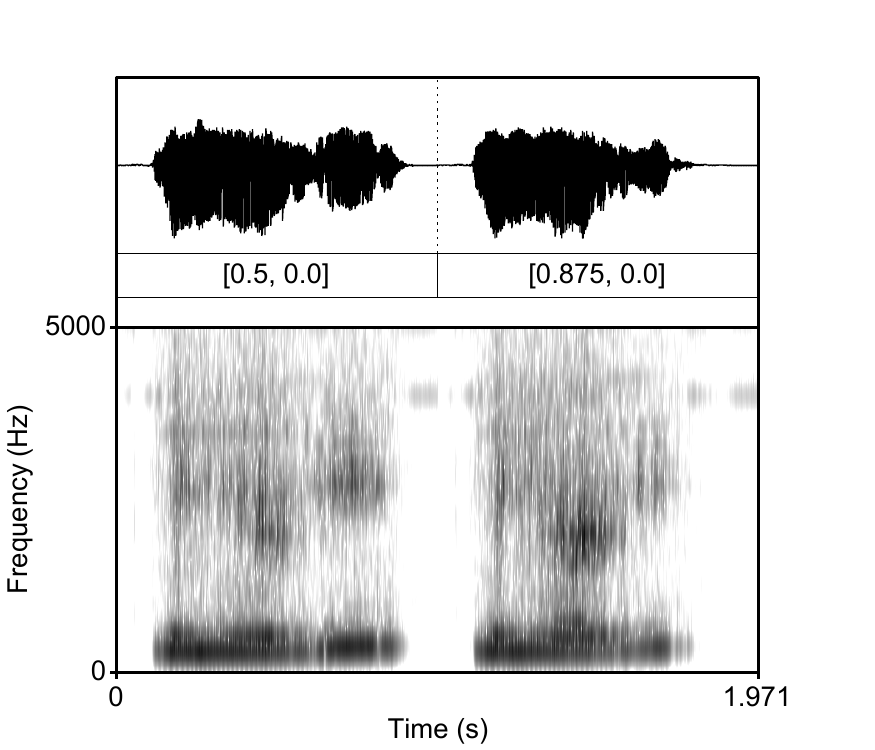}
\includegraphics[width=.19\textwidth]{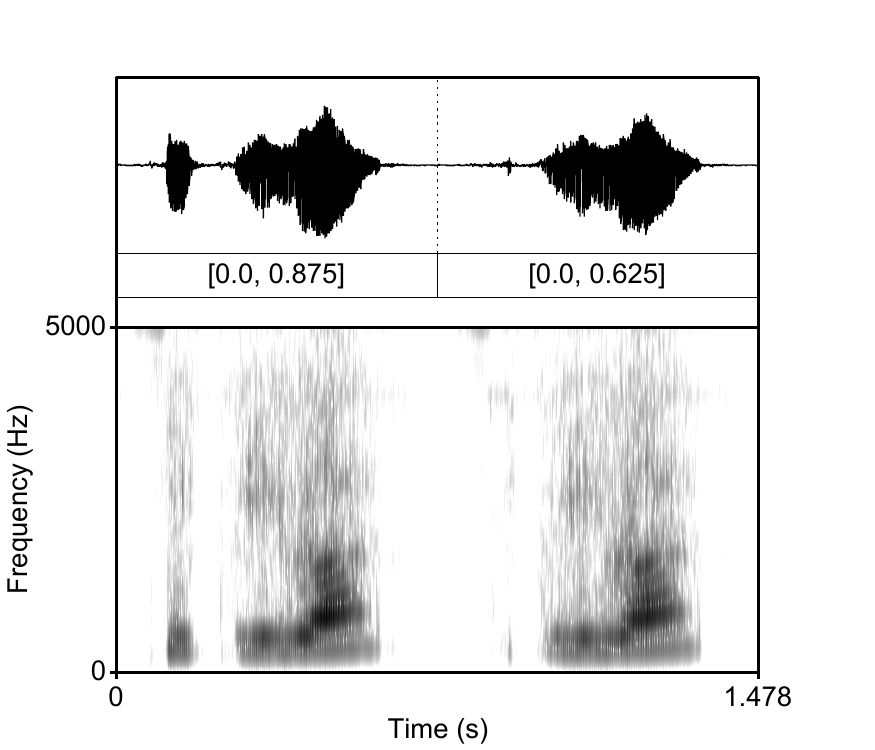}
\includegraphics[width=.19\textwidth]{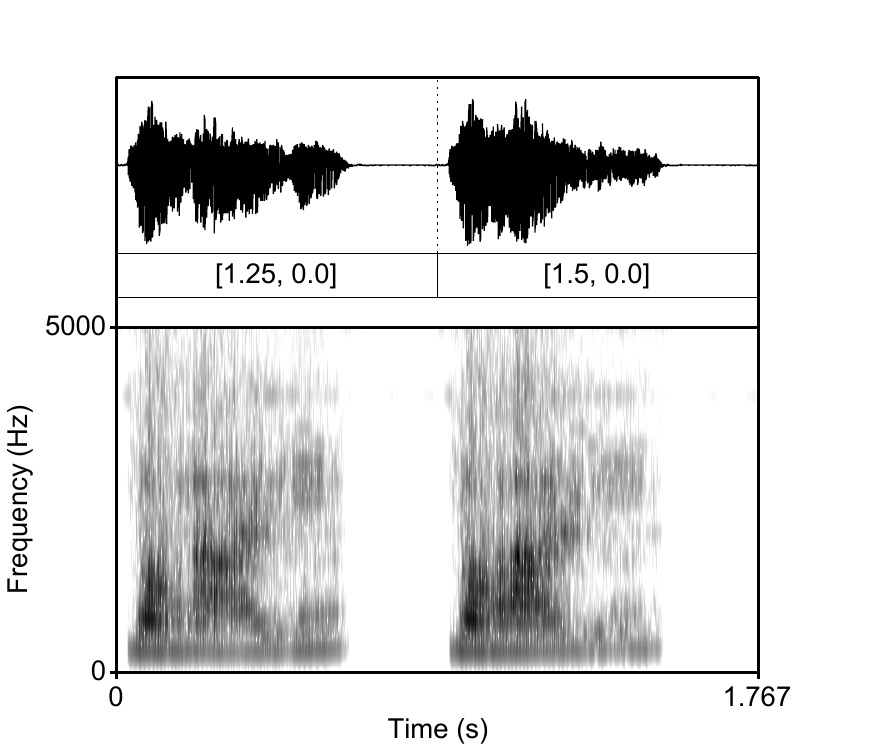}
\includegraphics[width=.19\textwidth]{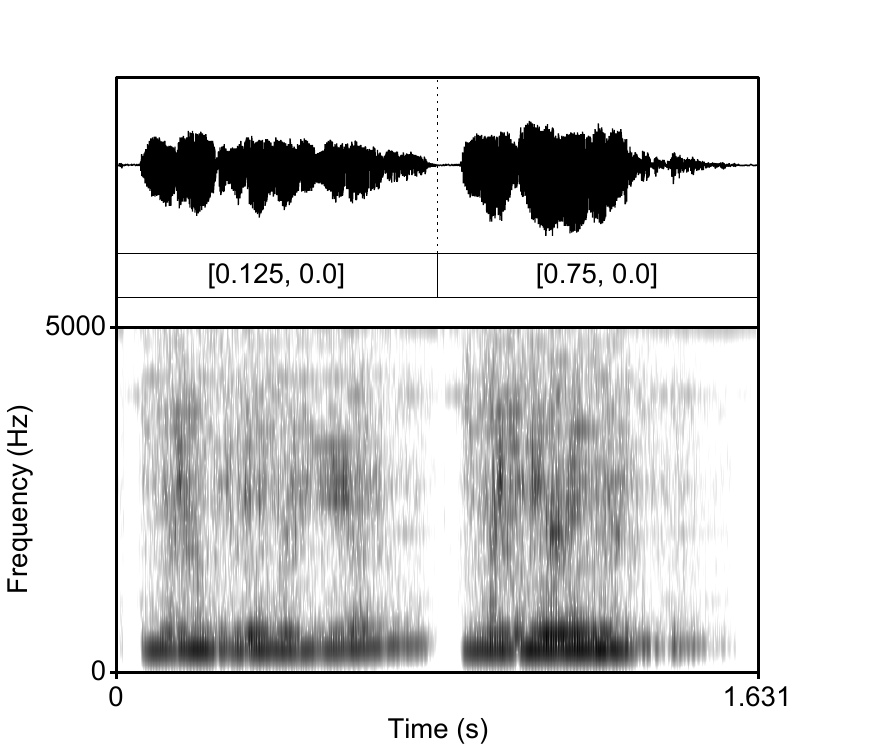}
\includegraphics[width=.19\textwidth]{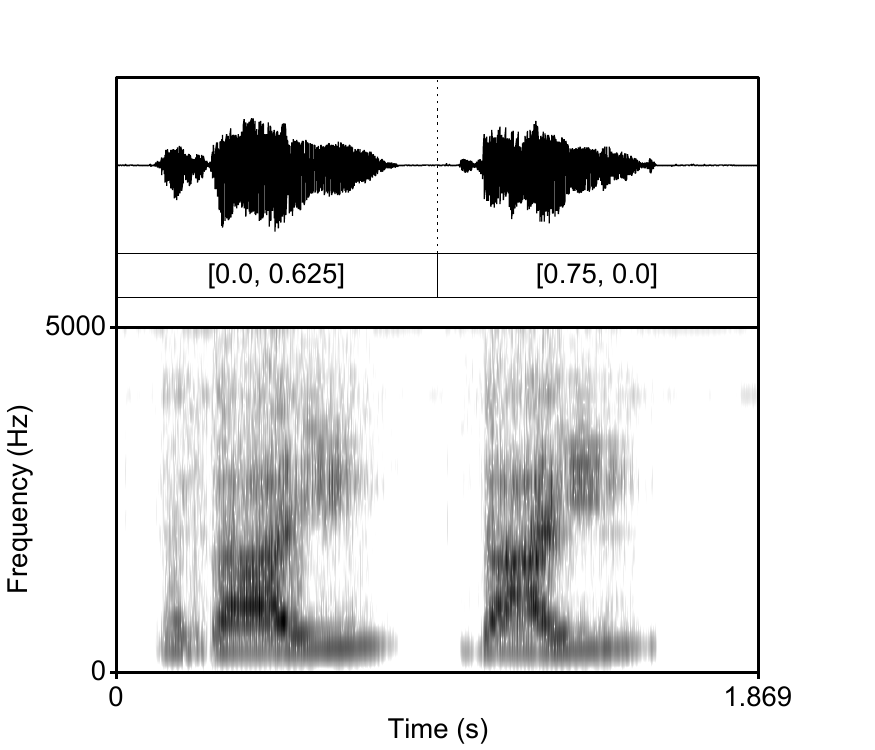}\\
\includegraphics[width=.19\textwidth]{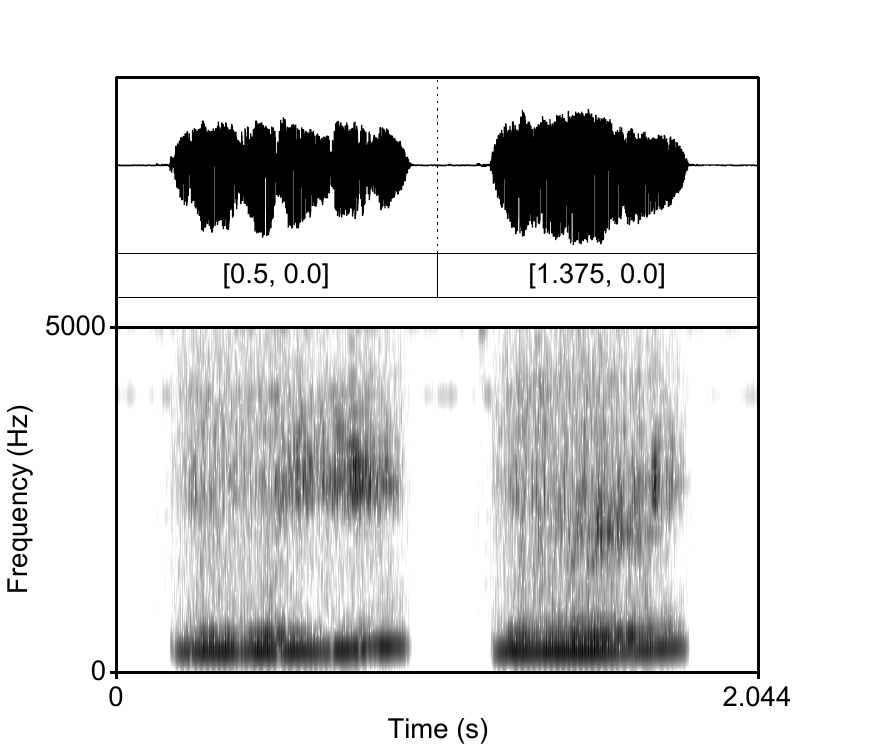}
\includegraphics[width=.19\textwidth]{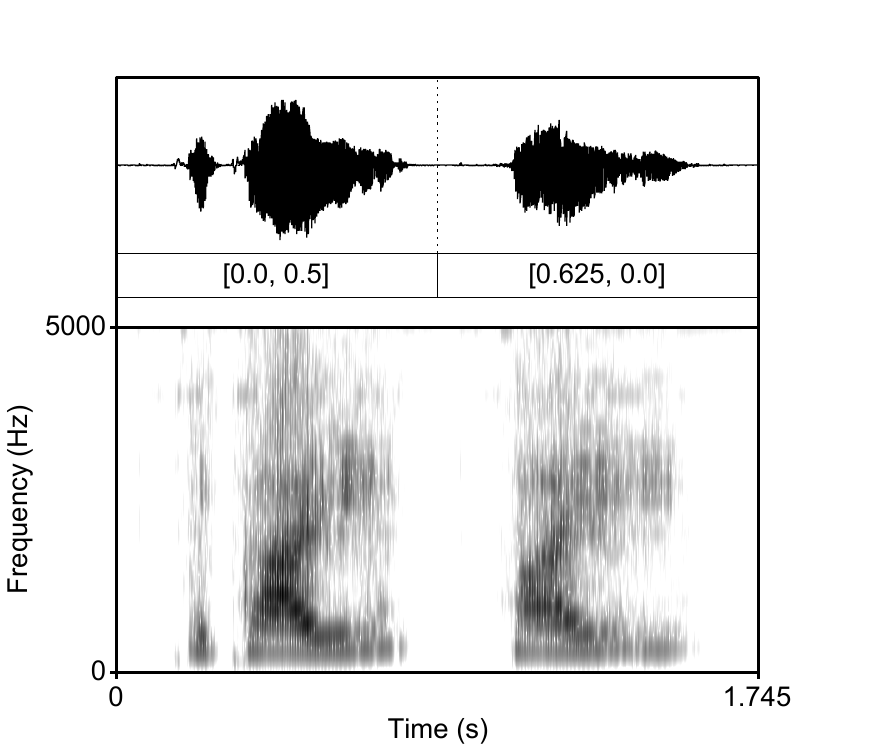}
\includegraphics[width=.19\textwidth]{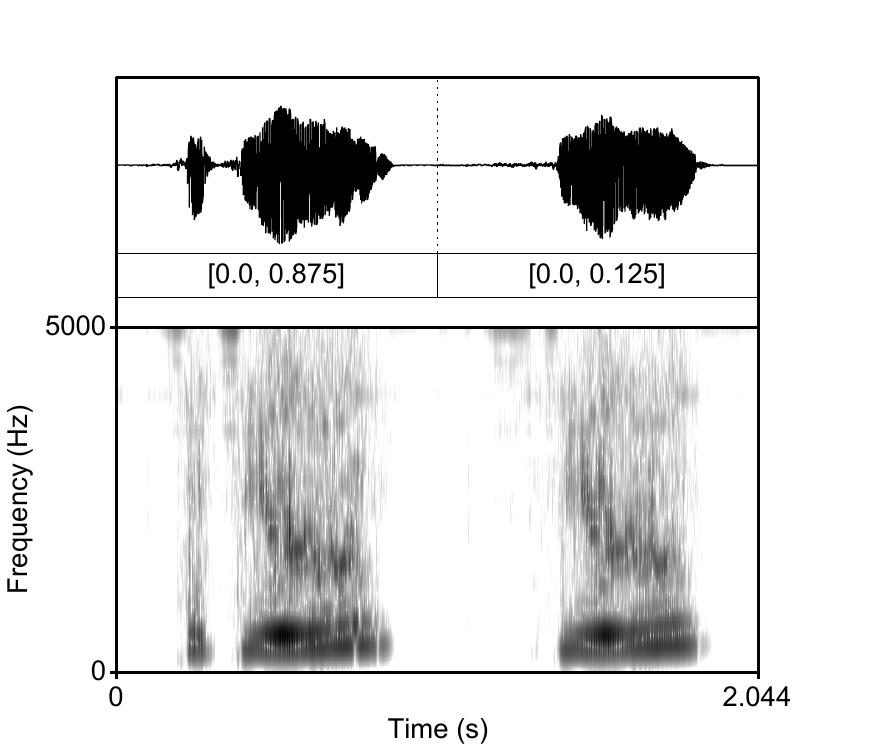}
\includegraphics[width=.19\textwidth]{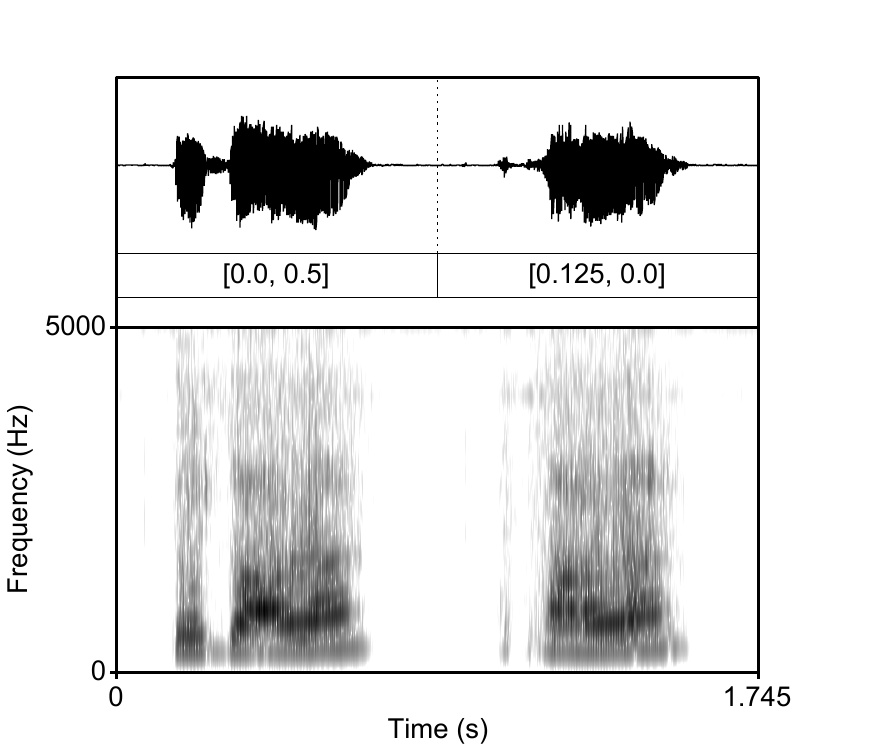}
\includegraphics[width=.19\textwidth]{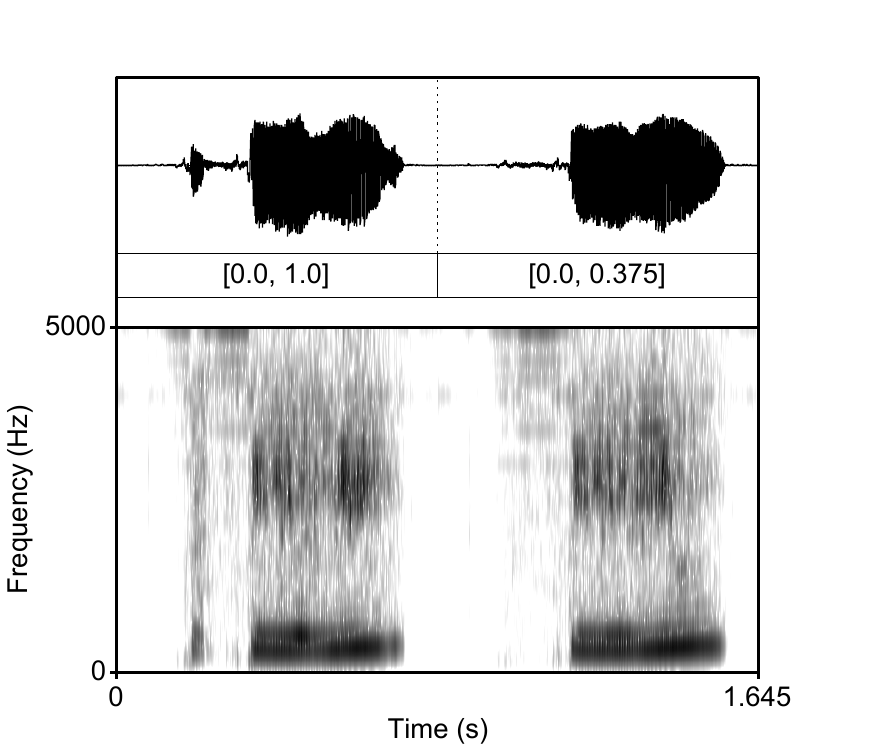}\\
\includegraphics[width=.19\textwidth]{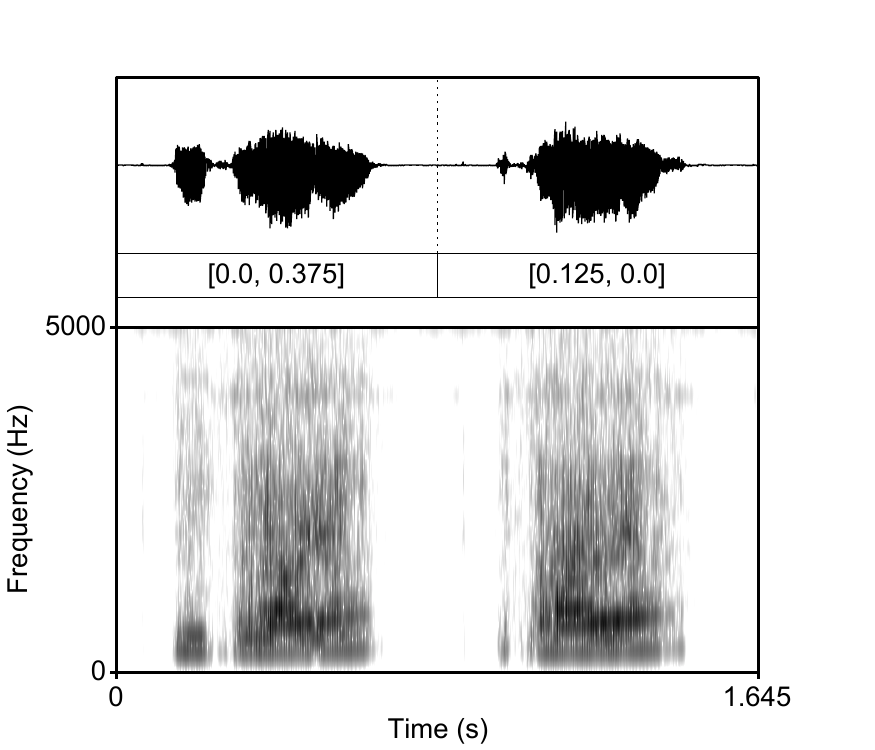}
\includegraphics[width=.19\textwidth]{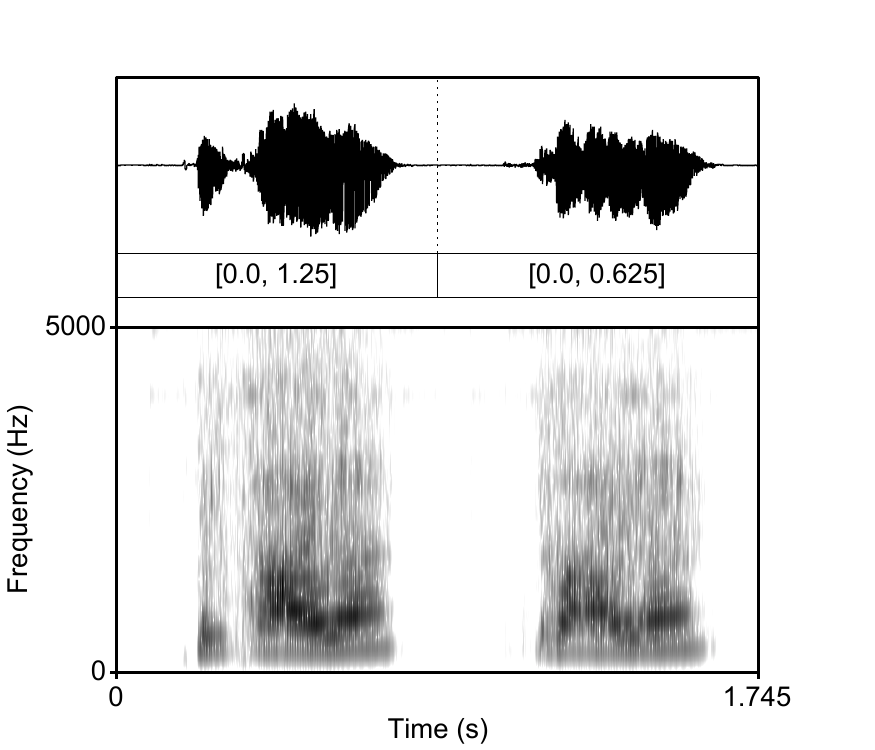}
\includegraphics[width=.19\textwidth]{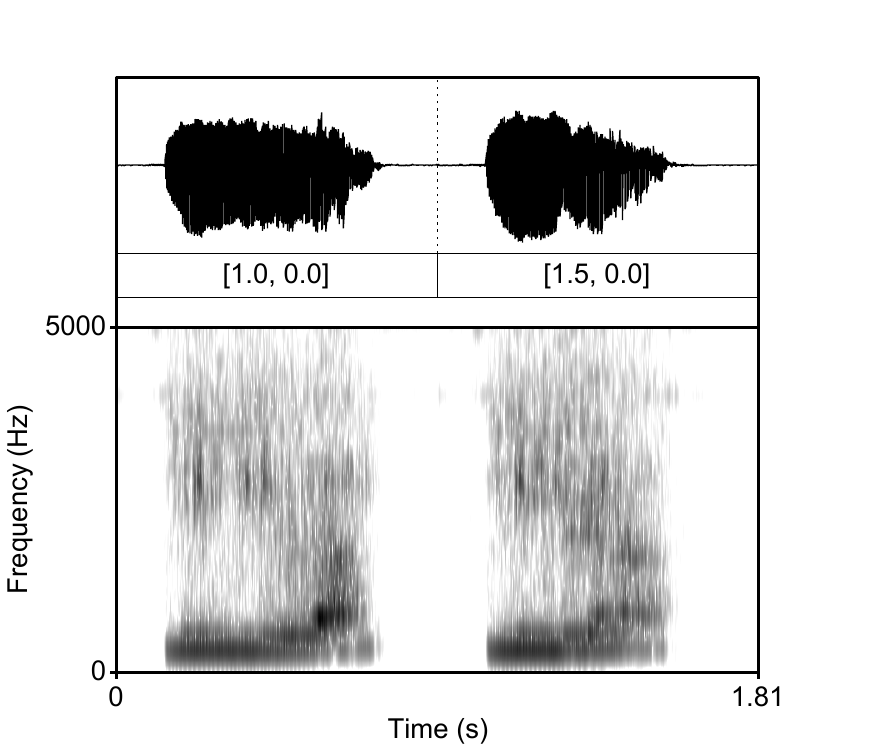}
\includegraphics[width=.19\textwidth]{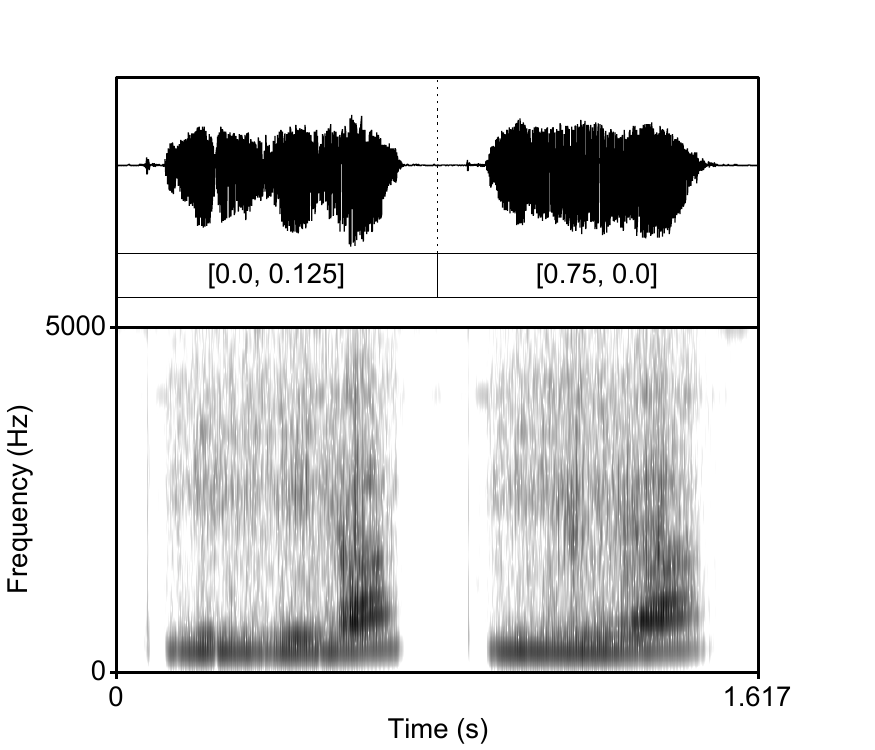}
\includegraphics[width=.19\textwidth]{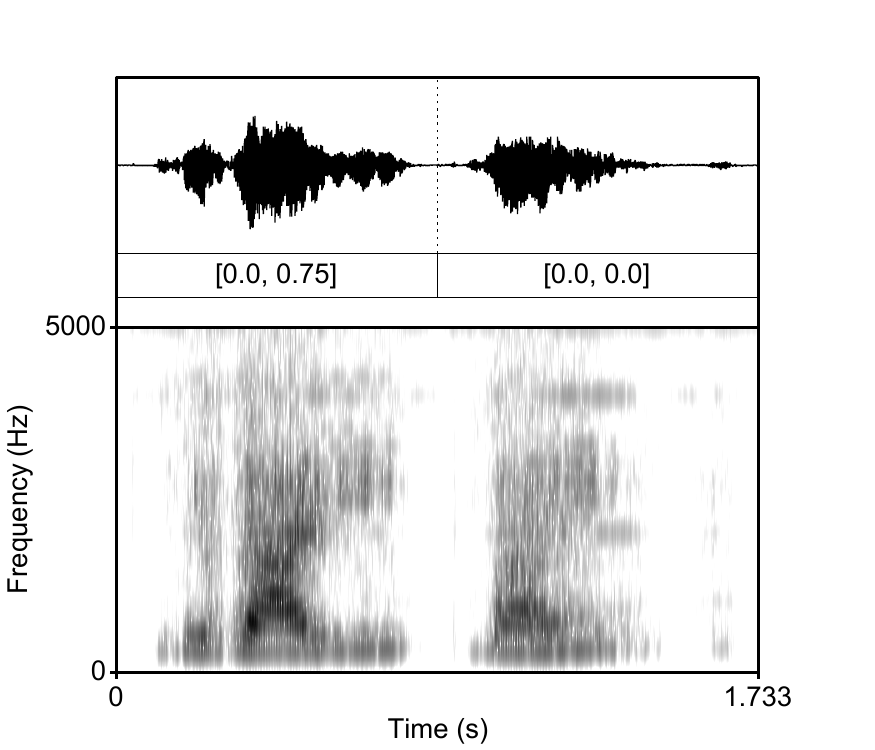}\\
\includegraphics[width=.19\textwidth]{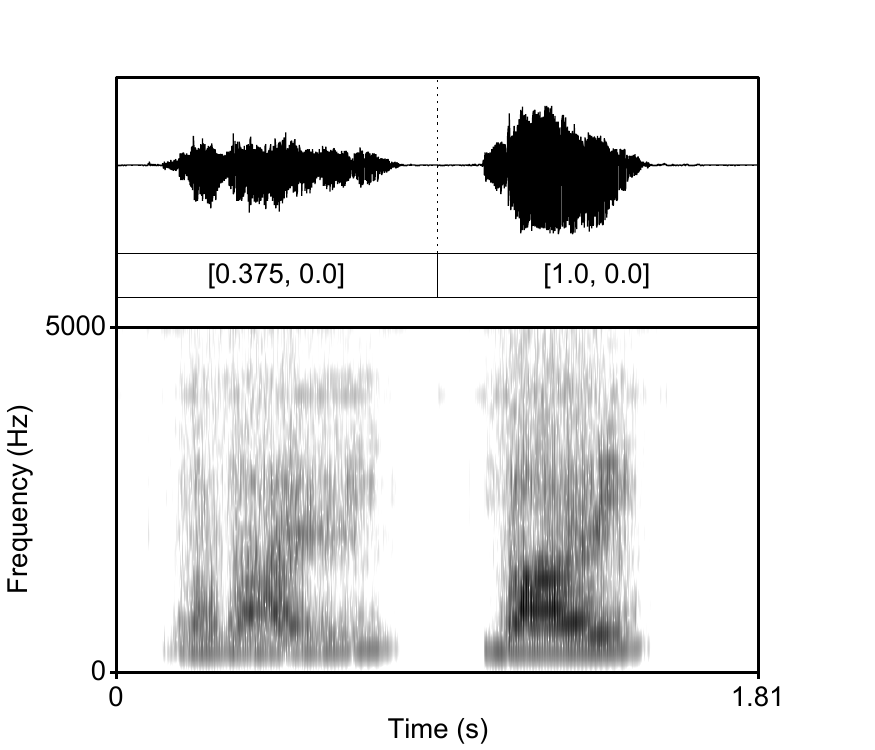}
\includegraphics[width=.19\textwidth]{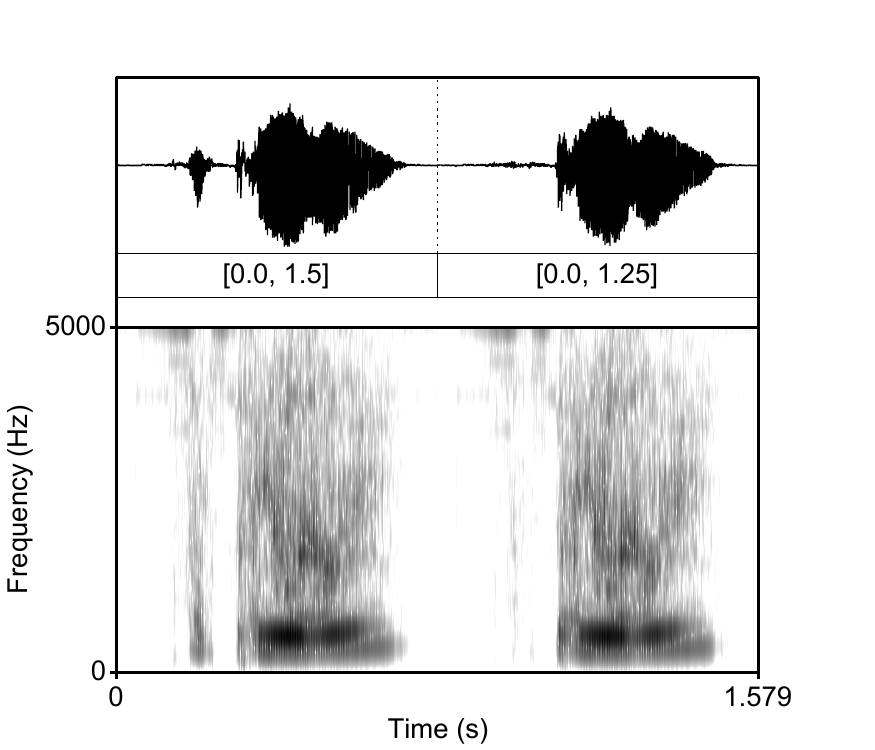}
\includegraphics[width=.19\textwidth]{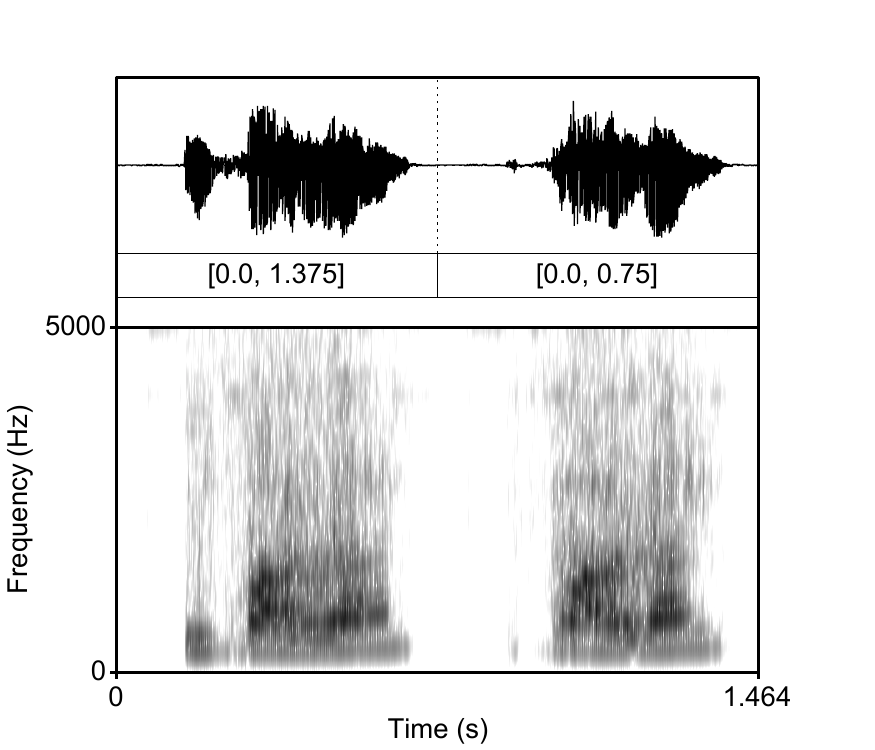}
\includegraphics[width=.19\textwidth]{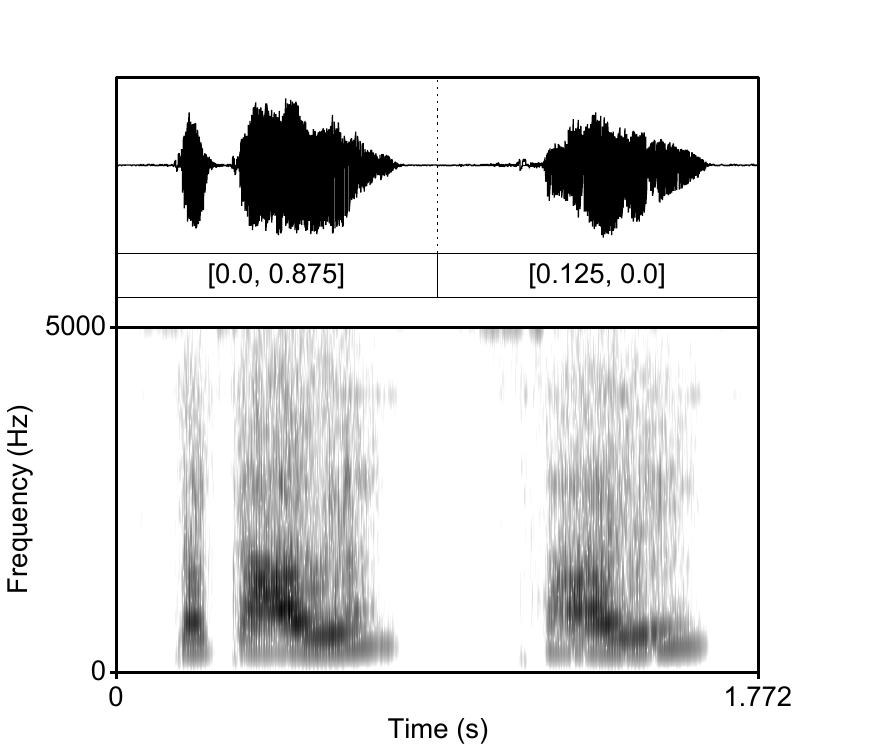}
\includegraphics[width=.19\textwidth]{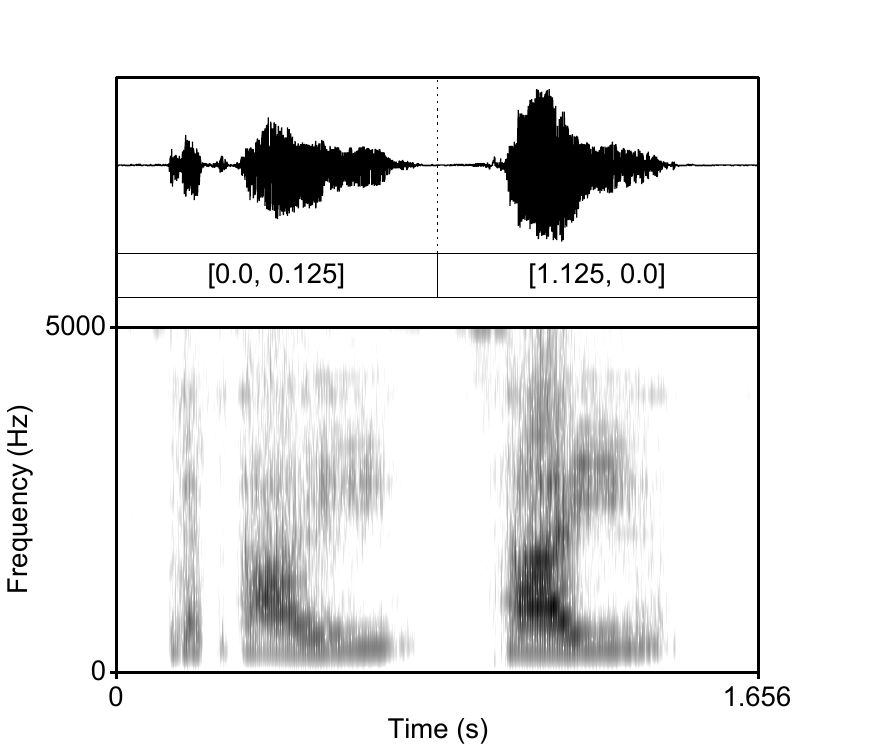}\\
\includegraphics[width=.19\textwidth]{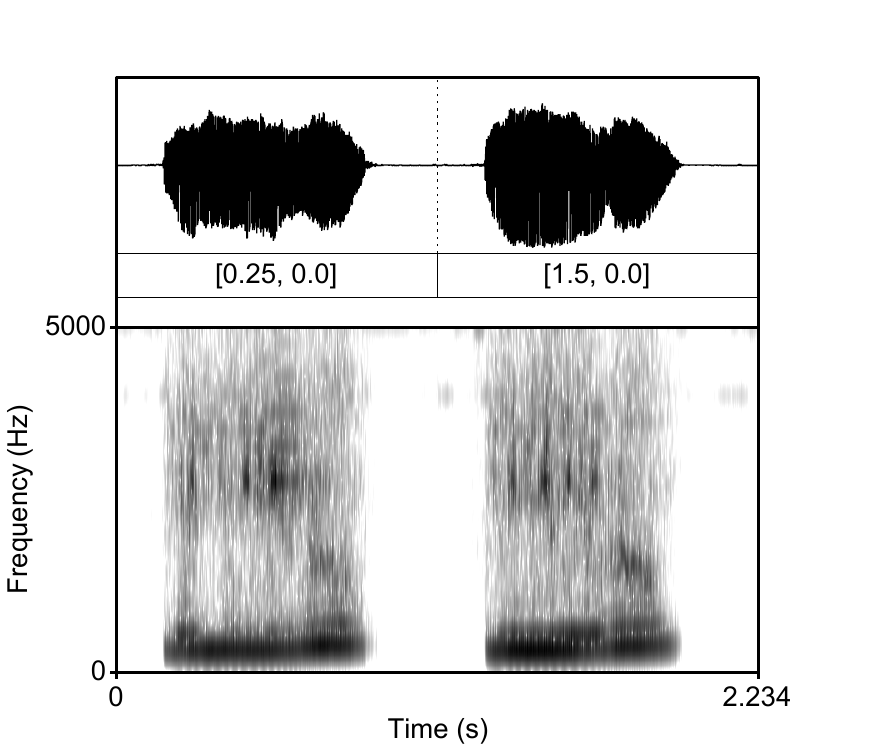}
\includegraphics[width=.19\textwidth]{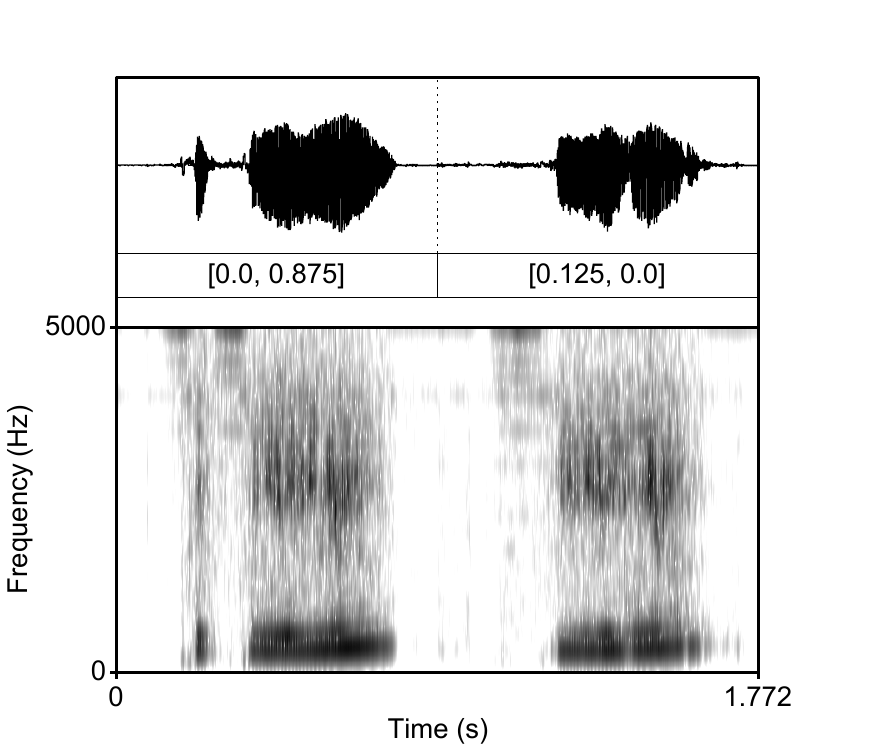}
\includegraphics[width=.19\textwidth]{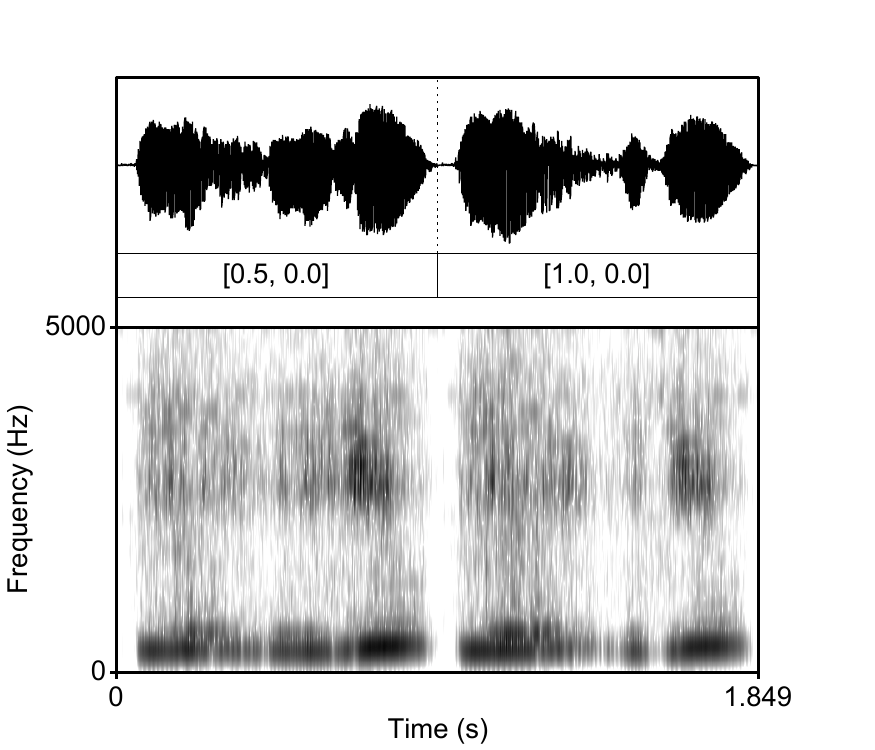}
\includegraphics[width=.19\textwidth]{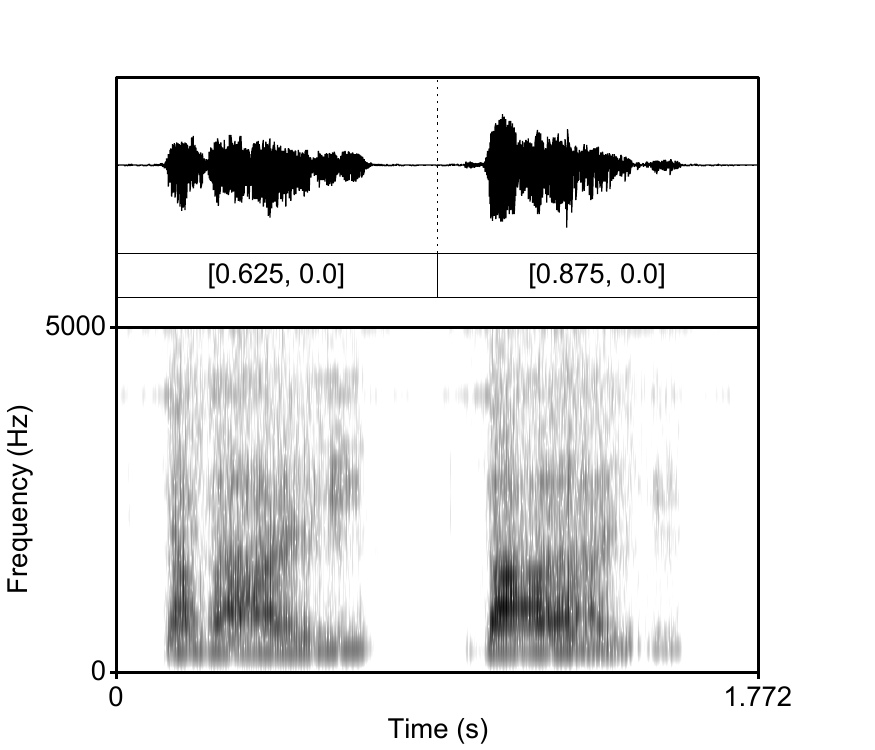}
\includegraphics[width=.19\textwidth]{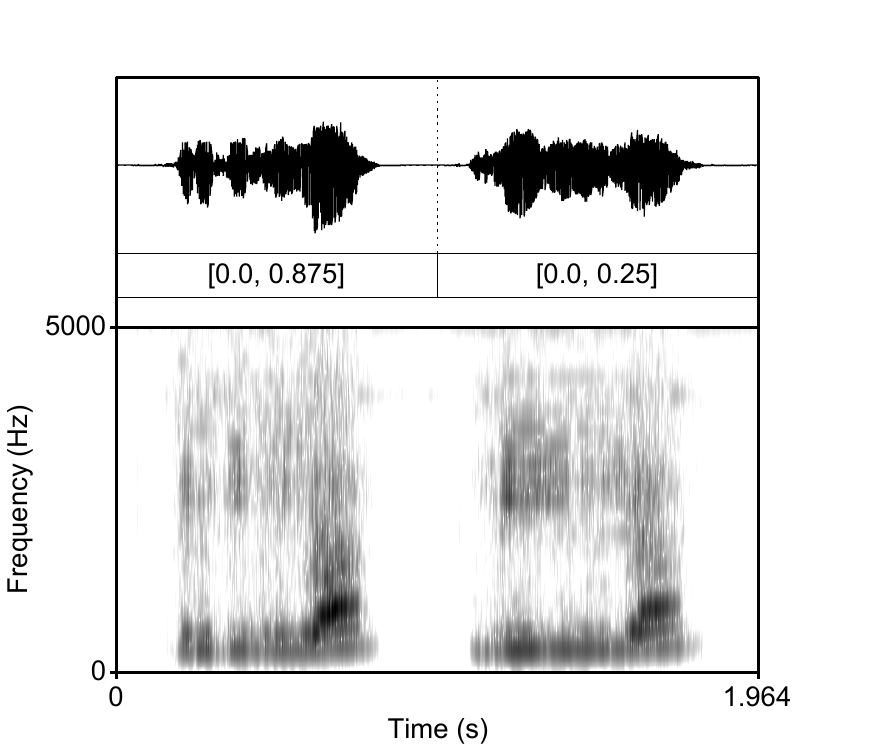}
\caption{Waveforms and spectrgrograms (0-5000 Hz) of the 25 reduplicated-unreduplicated form pairs annotated as involving no other major changes than the presence of the reduplicative syllable. Reduplicated outputs are on the left; unreduplicated outputs on the right. Values of the latent code are listed below each output; all other 98 $z$-variables are kept constant across the two pairs. In two cases involving a nasal C$_1$,  it is challenging to distinguish between very short remnants of the vocalic period of reduplication and periodic vibration of the nasal (the difference between reduplicated and unreduplicated form), but there is a clear contrast between the two forms of the pair.  \label{figures} 
}
\end{figure}

\end{document}